\documentclass[lettersize,journal]{IEEEtran}
\usepackage{amsmath,amsfonts}
\usepackage{algorithmic}
\usepackage{algorithm}
\usepackage{array}
\usepackage[caption=false,font=normalsize,labelfont=sf,textfont=sf]{subfig}
\usepackage{textcomp}
\usepackage{stfloats}
\usepackage{url}
\usepackage{verbatim}
\usepackage{graphicx}
\usepackage{cite}

\usepackage{bbding}
\usepackage{amssymb}
\usepackage{textcomp}
\usepackage[dvipsnames]{xcolor}
\usepackage{subfig}
\usepackage{tabularx}
\usepackage{makecell}
\usepackage{stfloats}
\usepackage{framed}
\usepackage{mydef}
\usepackage{multirow}
\usepackage[colorlinks,linkcolor=blue]{hyperref}
\def\BibTeX{{\rm B\kern-.05em{\sc i\kern-.025em b}\kern-.08em
    T\kern-.1667em\lower.7ex\hbox{E}\kern-.125emX}}

\newcommand{\stitle}[1]{\vspace{1ex} \noindent{\bf #1}}
\begin{document}

\title{A Survey of Large Language Models on Generative  Graph Analytics: Query, Learning, and Applications}


\author{Wenbo Shang and Xin Huang

\thanks{Wenbo Shang and Xin Huang are with the Department of Computer Science, Hong Kong Baptist University, Hong Kong, China (e-mail: cswbshang@comp.hkbu.edu.hk; xinhuang@comp.hkbu.edu.hk)}
}

\markboth{Journal of \LaTeX\ Class Files,~Vol.~14, No.~8, August~2021}%
{Shell \MakeLowercase{\textit{et al.}}: A Sample Article Using IEEEtran.cls for IEEE Journals}

\IEEEpubid{0000--0000/00\$00.00~\copyright~2021 IEEE}

\maketitle

\begin{abstract}
A graph is a fundamental data model to represent various entities and their complex relationships in society and nature, such as social networks, transportation networks, financial networks, and biomedical systems. Recently, large language models (LLMs) have showcased a strong generalization ability to handle {various natural language processing tasks} to answer users' arbitrary questions and {generate specific-domain content}. 
Compared with graph learning models, LLMs enjoy superior advantages in addressing the challenges of generalizing graph tasks by eliminating the need for training graph learning models and reducing the cost of manual annotation.
{However, LLMs are sequential models for textual data, but graphs are non-sequential topological data. It is challenging to adapt LLMs to tackle graph analytics tasks.} 
In this survey, we conduct a comprehensive investigation of existing LLM studies on graph data, which summarizes the relevant graph analytics tasks solved by advanced LLM models and points out the existing challenges and future directions.
Specifically, we study the key problems of LLM-based generative graph analytics (LLM-GGA) in terms of three categories:  
LLM-based graph query processing (LLM-GQP), LLM-based graph inference and learning (LLM-GIL), and graph-LLM-based applications. LLM-GQP focuses on an integration of graph analytics techniques and LLM prompts, including \emph{graph understanding} and \emph{knowledge graphs and LLMs}, while LLM-GIL focuses on learning and reasoning over graphs, including \emph{graph learning}, \emph{graph-formed reasoning} and \emph{graph representation}. 
We summarize the useful prompts incorporated into LLM to handle different graph downstream tasks. 
Moreover, we give a summary of LLM model evaluation, benchmark datasets/tasks, and a deep pro and cons analysis of the discussed LLM-GGA models. We also explore open problems and future directions in this exciting interdisciplinary research area of LLMs and graph analytics. 
\end{abstract}


\section{Introduction}

\begin{figure}
    \centering
     \includegraphics[width= 0.5\textwidth]{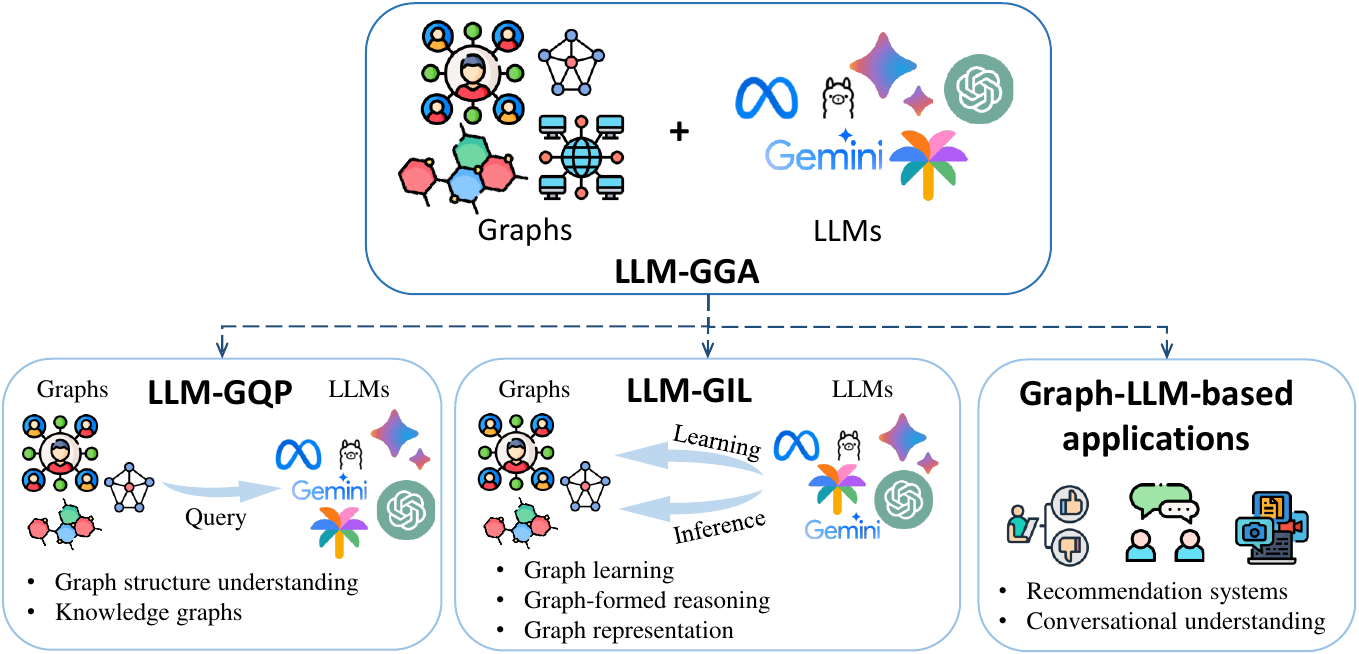}
      \caption{Illustration of the LLM-GGA domain. LLM-GGA domain includes three principal components: LLM-based graph query processing (LLM-GQP), which integrates graph analytics techniques and LLM prompts for query processing; LLM-based graph inference and learning (LLM-GIL), focusing on learning and reasoning over graphs; Graph-LLM-based applications that employ the graph-LLM framework to address non-graph tasks.}
      \label{introduction} 
\end{figure}

Large language models (LLMs) possess billions of parameters and have been trained on extensive corpora using training strategies like instruction tuning\cite{wei2022finetuned}\cite{liu2024visual} and Direct Preference Optimization (DPO)\cite{rafailov2024direct}, enabling them to exhibit powerful reasoning and semantic representation capabilities, thereby advancing AI intelligence closer to human levels. Undoubtedly, LLMs currently serve as the foundation model for NLP tasks\cite{huang2024trustllm}\cite{touvron2023llama}\cite{ouyang2022training}, showcasing strong generalization abilities to handle various NLP tasks such as question answering\cite{sun2025causalabstain}\cite{zhuang2023toolqa}\cite{li2024flexkbqa}, machine translation\cite{zhang2023prompting}, code generation\cite{liu2024your}\cite{ni2023lever}, etc. LLMs have demonstrated extensive common knowledge and robust semantic comprehension abilities, fundamentally transforming existing text-processing workflows. {While initially designed for textual data, LLMs are increasingly being utilized for tasks beyond language processing, particularly graph analytic tasks, aiming to leverage the robust and strong capabilities of LLMs across different tasks to achieve superior performance.}

Graphs, as structured data, play a crucial role in various real-world application scenarios, including citation networks \cite{sen2008collective}, social networks \cite{hamilton2017inductive}, molecular graphs \cite{wu2018moleculenet}, web links \cite{broder2000graph}, and to name a few. 
Various graph analytics tasks have been studied to show their usefulness, e.g., node classification, link prediction, subgraph mining, influence maximization, and so on.   
Their versatility and ability to capture complex relationships have made graphs indispensable tools in academic research and industry platforms. Recently, one kind of graph-based learning model, graph neural network (GNN)\cite{kipf2016semi}\cite{velickovic2017graph}, has been widely studied and applied to solve challenging graph tasks. GNNs utilize recursive message passing\cite{gilmer2017neural} and aggregation mechanisms\cite{hong2009aggregation} among nodes to derive representations of nodes, edges, or entire graphs, capturing both graph structure and node features, which have been used for various downstream tasks. 
{However, GNNs exhibit weak generalization capabilities and semantic representation abilities\cite{cong2021provable}\cite{fan2023generalizing}\cite{li2024survey}, requiring retraining for different specific graph tasks and showing limited transferability\cite{ren2024survey}\cite{liu2024one}. }
\IEEEpubidadjcol

\begin{table*}[h!t]
\scriptsize
   \centering
   \renewcommand\arraystretch{1.15}
   \setlength\tabcolsep{2.5pt}
   \caption{Comparison of LLM-GGA and other surveys.}
   \begin{tabular}{l|c|c|c|c|c|c|c}\hline
    Surveys &  Graph structure understanding & KGs and LLMs &  Graph learning &  Graph-formed reasoning & Graph representation &   Graph-LLM-based applications & Year \\ \hline
    Li et al.\cite{li2024survey} & \XSolidBrush & \XSolidBrush &  \Checkmark & \XSolidBrush & \Checkmark  & \XSolidBrush & 2024 \\
    Jin et al.\cite{jin2024large} & \XSolidBrush & \XSolidBrush &  \Checkmark & \XSolidBrush & \Checkmark  & \XSolidBrush & 2024 \\
    Liu et al.\cite{liu2025graph} & \XSolidBrush & \XSolidBrush &  \Checkmark & \XSolidBrush & \XSolidBrush  & \XSolidBrush & 2024 \\
    Ren et al.\cite{ren2024survey} & \XSolidBrush & \XSolidBrush &  \Checkmark & \XSolidBrush & \Checkmark  & \XSolidBrush & 2024 \\
    Li et al.\cite{li2024graph} & \XSolidBrush & \XSolidBrush &  \Checkmark & \XSolidBrush & \XSolidBrush  & \XSolidBrush & 2024 \\
    Peng et al.\cite{peng2024graph} & \XSolidBrush & \Checkmark &  \XSolidBrush & \XSolidBrush & \XSolidBrush  & \XSolidBrush & 2024 \\
    Han et al.\cite{han2024retrieval} & \XSolidBrush & \Checkmark &  \XSolidBrush & \XSolidBrush & \XSolidBrush  & \XSolidBrush & 2024 \\
    Zhang et al.\cite{zhang2025survey} & \XSolidBrush & \Checkmark &  \XSolidBrush & \XSolidBrush & \XSolidBrush  & \XSolidBrush & 2024 \\
    Lavrinovics et al.\cite{lavrinovics2025knowledge} & \XSolidBrush & \Checkmark &  \XSolidBrush & \XSolidBrush & \XSolidBrush  & \XSolidBrush & 2024 \\

    \hline
    \textbf{LLM-GGA} & \Checkmark & \Checkmark & \Checkmark & \Checkmark & \Checkmark  & \Checkmark & 2025\\
    \hline
    \end{tabular}
    \label{summary-survey}
\end{table*}

{Besides a simple graph model associated with no attributes or category/numerical attributes,  nodes may be enriched with raw and detailed text attributes, capturing rich and comprehensive semantics, known as \emph{text-attributed graphs}. For instance, in citation networks, nodes represent papers and edges denote citations between them, with the nodes containing long text attributes such as titles, keywords, abstracts, and even the full articles~\cite{he2023harnessing}\cite{yan2023comprehensive}\cite{zhao2023learning}\cite{fang2024gaugllm}. In healthcare networks, nodes link patients, doctors, diseases, and treatments, incorporating the text attributes of medical records to enhance patient care~\cite{tan2024walklm}\cite{choi2017gram}\cite{li2022graph}. Similarly, in social networks, nodes represent users and edges represent interactions, with nodes containing text attributes like user bios and post content to analyze social influence~\cite{liao2018attributed}\cite{campbell2013social}\cite{li2022distilling}. In the past day, analyzing these long text attributes has been challenging due to the limited ability of previous language models\cite{he2023harnessing}. Leveraging the advanced capabilities of recent LLMs, it is promising to explore and conduct text-attributed graph analytics.} 

{Therefore, adapting LLMs' powerful reasoning, semantic representation, and generalization capabilities to address graph tasks, leading to the inspiration of a graph foundation model\cite{liu2025graph}\cite{jin2024large}\cite{ren2024survey}, is the core of current efforts in leveraging existing large language models for various graph-related tasks. However, LLMs are sequential models for language processing,  while graphs are non-Euclidean topological data, which brings a key challenge:
\textit{\textbf{how can LLMs solve graph data tasks?}} 
More specifically, we study this core challenge by answering three detailed questions: (a) What specific graph tasks can LLMs answer? (b) How do LLMs tackle these tasks? (c) What is the effectiveness of LLM-based methods in solving these tasks compared with the existing graph-based approaches?} 

To address the above question, this survey conducts a comprehensive study of existing relevant work on graph analytics and LLMs, focusing on exploring the key issue of the LLM-based generative graph analytics (LLM-GGA) field. Drawing from a thorough investigation of the LLM-GGA domain, we offer a structured and methodical analysis that delineates the field into three principal components: LLM-based graph query processing (LLM-GQP), which necessitates the melding of graph analytics techniques and LLM prompts for query processing; LLM-based graph inference and learning (LLM-GIL), focusing on learning and reasoning over graphs; and lastly, graph-LLM-based applications that employ the graph-LLM framework to address non-graph tasks, such as recommendation systems. The framework is shown in Figure~\ref{introduction}.

We categorize these three main components into a total of six directions to provide a guideline for researchers to conduct more in-depth studies. LLM-GQP includes graph understanding, and knowledge graphs and LLMs directions. LLM-GIL covers graph learning, graph-formed reasoning, and graph representation directions. The sixth direction is graph-LLM-based applications. The following section details these six directions:
\begin{itemize}
    \item \textbf{Graph understanding tasks.} This research direction studies the performance of LLMs over P problems and NP-hard problems, {exploring whether LLMs can comprehend graph structures to conduct graph algorithmic tasks and graph structural properties.} Current methods have primarily explored LLMs' understanding of graph structures, such as shortest path, clustering coefficient computation\cite{guo2023gpt4graph}\cite{liu2023evaluating}, and more complex problems like maximum flow and Hamilton path\cite{zhang2024llm4dyg}\cite{huang2024can}\cite{wang2024can}. Furthermore, many more NP-hard problems have been explored, such as maximum clique, graph coloring and the proof of $P\neq NP$ \cite{dong2023large}\cite{fan2023nphardeval}. 
    Two main methods are introduced: prompting and supervised fine-tuning (SFT). The prompting methods explore the LLM's current structural understanding ability through query processing. Meanwhile, SFT methods enhance LLMs' structure understanding capability by tuning them on specific graph datasets.
    \item {\textbf{Knowledge graphs and LLMs.} This direction investigates the relationship between LLMs and Knowledge Graphs (KGs). With the emergence of LLMs, discussions have arisen regarding the potential replacement of KGs\cite{dernbach2024glam}\cite{sun2024think}\cite{tian2024graph}\cite{luo2024reasoning}. Consequently, this paper discusses the limitations of LLMs in processing factual knowledge, evaluates strategies for improving LLM efficacy via KGs, and investigates potential avenues for future advancements.}
    \item \textbf{Graph learning tasks.} {This direction explores whether LLMs can combine graph structure and text attributes for learning, extracting high-dimensional features of nodes, edges, and graphs from embedding spaces, and understanding the semantic information of graphs, for example, tasks like node classification, link prediction and graph construction\cite{he2023harnessing}\cite{zhao2023learning}\cite{xiong2024large}\cite{liu2024one}}. Most methods can be categorized into three methodologies: LLMs-as-enhancers, LLMs-as-predictors and LLMs-as-graph-generators. LLMs can leverage their powerful reasoning ability and vast knowledge base to enhance GNNs, predict results directly and generate graphs from natural language.
    \item \textbf{Graph-formed reasoning.} This direction explores how LLMs utilize graph structures to simulate human thinking patterns during the reasoning process \cite{besta2024graph}\cite{zhang2023cumulative}\cite{yao2023beyond}, enabling them to solve more complex reasoning problems such as algorithmic, logical, and mathematical tasks. Graph-formed reasoning involves two types of reasoning: think on the graph and verify on the graph. \emph{Think on the graph} refers to LLMs deriving the final conclusion through the graph-like reasoning. \emph{Verify on the graph} refers to verifying the correctness of the LLMs' intermediate or final outputs through the graph structure.
    \item \textbf{Graph representation.} This direction explores enhancing graph representation with LLMs, particularly for Text Attribute Graphs (TAGs). LLMs' strong text representation capabilities allow text embeddings to capture deeper semantic nuances. However, the key challenge in this area remains how to capture and integrate graph structure into graph representation effectively\cite{zhao2024graphtext}\cite{tan2024walklm}\cite{qin2023disentangled}\cite{shang2024path}. There are three forms of graph representation: graph embedding, graph-enhanced text embedding, and graph-encoded prompts. Graph embedding methods transform a graph into a sequential format for LLM processing. Graph-enhanced text embedding methods integrate structure into text embedding, where the integration method can be concatenated. Graph-encoded prompts focus on the way a graph is described within prompts. Lastly, we conduct a technical comparison and analysis of supervised and unsupervised graph representation learning methodologies, aiming to guide readers in their future research endeavors.

    \item \textbf{Graph-LLM-based applications.} This part explores the tasks where graph-LLM-based methods can be applied for useful downstream application\cite{wei2024llmrec}\cite{wang2023graph}\cite{wu2024exploring}, such as task planning and recommendation systems.
\end{itemize}
\tikzstyle{leaf}=[draw=hiddendraw,
    rounded corners,
    minimum height=1em,
    fill=mygreen!40,
    text opacity=1, 
    align=center,
    fill opacity=.5,  
    text=black,
    align=left,
    font=\scriptsize,
    inner xsep=3pt,
    inner ysep=1pt,
    ]
\tikzstyle{middle}=[draw=hiddendraw,
    rounded corners,
    minimum height=1em,
    fill=output-white!40, 
    text opacity=1, 
    align=center,
    fill opacity=.5,  
    text=black,
    align=left,
    font=\scriptsize,
    inner xsep=3pt,
    inner ysep=1pt,
    ]
\begin{figure*}[ht]
\centering
\begin{forest}
  for tree={
  forked edges,
  grow=east,
  reversed=true,
  anchor=base west,
  parent anchor=east,
  child anchor=west,
  base=middle,
  font=\scriptsize,
  rectangle,
  line width=0.7pt,
  draw=output-black,
  rounded corners,align=left,
  minimum width=2em,
    s sep=2.5pt,
    inner xsep=3pt,
    inner ysep=1pt,
  },
  where level=1{text width=4.5em}{},
  where level=2{text width=6em,font=\scriptsize}{},
  where level=3{font=\scriptsize}{},
  where level=4{font=\scriptsize}{},
  where level=5{font=\scriptsize}{},
  [LLM-GGA ,middle,rotate=90,anchor=north,edge=output-black
    [Graph Understanding, middle, edge=output-black, text width=8.7em
        [Prompting LLMs, middle, text width=9.5em, edge=output-black
            [Manual prompts, middle, text width=5em, edge=output-black 
                [NLGraph \cite{wang2024can}{,}  LLM4DyG \cite{zhang2024llm4dyg}{,} GraphArena \cite{tang2025grapharena}{,} NLGIFT \cite{zhang2024can}{,}\\ GraphInsight \cite{cao2024graphinsight}{,} LLMtoGraph \cite{liu2023evaluating}{,}  NPHardEval \cite{fan2023nphardeval} and\\ other works \cite{huang2024can}{,} \cite{dong2023large}   , leaf, text width=18.6em, edge=output-black]
            ]
            [Self-prompting, middle, text width=5em, edge=output-black 
                [GPT4Graph \cite{guo2023gpt4graph}{,} TalkGraph \cite{fatemi2024talk}, leaf, text width=18.6em, edge=output-black]
            ]
            [API call prompts, middle, text width=5em, edge=output-black 
                [Gorilla\cite{patil2024gorilla}{,} Graph-ToolFormer \cite{zhang2023graph}, leaf, text width=18.6em, edge=output-black]
            ]
        ]
        [Supervised fine-tuning LLMs, middle, text width=9.5em, edge=output-black
            [GraphLLM\cite{chai2023graphllm} , leaf, text width=5em, edge=output-black]
        ]
    ]
    [Graph Learning, middle, edge=output-black, text width=8.7em
        [LLMs-as-enhancers, middle, text width=9.5em, edge=output-black
            [Encoding text attributes into embeddings, middle, text width=12.2em, edge=output-black
                [TAPE \cite{he2023harnessing}{,}  GraphAdapter \cite{huang2024can2}{,} GLEM \cite{zhao2023learning}{,}  \\ OFA \cite{liu2024one}{,}SIMTEG~\cite{duan2023simteg}{,}GAugLLM \cite{fang2024gaugllm} ,   leaf, text width=12.6em, edge=output-black]
            ]
            [Generating pseudo labels, middle, text width=12.2em, edge=output-black
                [GLEM \cite{zhao2023learning}{,} LLM-GNN~\cite{chen2024label} , leaf, text width=12.6em, edge=output-black]
            ]
            [Providing external knowledge/explanations, middle, text width=12.2em, edge=output-black
                [Graph-LLM~\cite{chen2024exploring}{,} TAPE~\cite{he2023harnessing} , leaf, text width=12.6em, edge=output-black]
            ]
        ]
        [LLMs-as-predictors, middle, text width=9.5em, edge=output-black
            [Prompting LLMs, middle, text width=9em, edge=output-black
                [Beyond Text~\cite{hu2023beyond}{,} KG-LLM~\cite{shu2024knowledge}{,} LPNL~\cite{bi2024lpnl} , leaf, text width=14.5em, edge=output-black]
            ]
            [Fine-tuning LLMs, middle, text width=9em, edge=output-black
                [IntructGLM~\cite{ye2024language}{,} GraphLLM \cite{chai2023graphllm} , leaf, text width=14.5em, edge=output-black]
            ]
            [Efficient tunable modules, middle, text width=9em, edge=output-black
                [LLaGA~\cite{chen2024llaga}{,} GraphTranslator~\cite{zhang2024graphtranslator}{,} ENGINE~\cite{zhu2024efficient} , leaf, text width=14.5em, edge=output-black]
            ]
        ]
        [LLMs-as-graph-generator, middle, text width=9.5em, edge=output-black
            [PiVe~\cite{han2024pive}{,} TG-LLM~\cite{xiong2024large}{,} EDC~\cite{zhang2024extract} , leaf, text width=25.5em, edge=output-black]
        ]
    ]
    [Graph-formed reasoning, middle, edge=output-black, text width=8.7em
        [Think on the graph, middle, text width=9.5em, edge=output-black
            [GoT* \cite{yao2023beyond}{,} GoT \cite{besta2024graph}{,} MindMap~\cite{wen2024mindmap}{,} CR~\cite{zhang2023cumulative} and other works \cite{vashishtha2023causal}, leaf, text width=25.5em, edge=output-black]
        ]
        [Verify on the graph, middle, text width=9.5em, edge=output-black
            [GraphReason~\cite{cao2023enhancing}{,}   Graph-guided CoT~\cite{park2023graph}{,} D-UE~\cite{da2024llm},leaf, text width= 25.5em, edge=output-black]
        ]
    ]
    [Graph representation, middle, edge=output-black, text width=8.7em
        [Graph embedding, middle, text width=9.5em, edge=output-black
            [WalkLM~\cite{tan2024walklm}{,} Path-LLM\cite{shang2024path}{,} GraphText~\cite{zhao2024graphtext}{,} GALLON~\cite{xu2024llm}, leaf, text width=25.5em, edge=output-black]
        ]
        [Graph-enhanced text embedding, middle, text width=9.5em, edge=output-black
            [DGTL~\cite{qin2023disentangled}{,}  G2P2~\cite{wen2023augmenting},leaf, text width= 25.5em, edge=output-black]
        ]
        [Graph-encoded prompts, middle, text width=9.5em, edge=output-black
            [TalkGraph~\cite{fatemi2024talk}{,} GraphTMI~\cite{das2024modality},leaf, text width= 25.5em, edge=output-black]
        ]
    ]
    [Knowledge graphs and LLMs, middle, edge=output-black, text width=8.7em
        [KGs tackle LLM limitations, middle, text width=9.5em, edge=output-black
            [Head-to-Tail~\cite{sun2024head}{,} G-retriever~\cite{he2024g}{,} ToG~\cite{sun2024think}{,} RoG~\cite{luo2024reasoning}{,} GLaM~\cite{dernbach2024glam}{,} KGPLMs~\cite{yang2024give}{,}\\ GNP~\cite{tian2024graph}{,} GraphRAG~\cite{edge2024local}{,}  ToG-2~\cite{ma2024think}, leaf, text width=25.5em, edge=output-black]
        ]
        [LLMs tackle KG limitations, middle, text width=9.5em, edge=output-black
            [AutoAlign~\cite{zhang2023autoalign}{,} EtD~\cite{liu2024explore}{,}  MuKDC~\cite{li2024llm}{,} KGP\cite{wang2024knowledge} and other works \cite{shah2024improving}{,} \cite{sehwag2024context}{,} \cite{giarelis2024unified}, leaf, text width=25.5em, edge=output-black]
        ]
    ]
    [Graph-LLM-based application, middle, edge=output-black, text width=8.7em
        [Task planning, middle, text width=9.5em, edge=output-black
            [Optimus-1~\cite{li2024optimus}{,} SayPlan~\cite{rana2023sayplan}, leaf, text width=25.5em, edge=output-black]
        ]
        [Recommendation system, middle, text width=9.5em, edge=output-black
            [LLMHG~\cite{chu2024llm} and other works ~\cite{abu2024knowledge} ,leaf, text width= 25.5em, edge=output-black]
        ]
        [other application, middle, text width=9.5em, edge=output-black
            [MANAGER~\cite{ouyang2024modal}{,}  GRACE~\cite{lu2024grace}{,} GPT4GNAS~\cite{wang2023graph} and other works \cite{li2024llm2},leaf, text width= 25.5em, edge=output-black]
        ]
    ]
  ]
\end{forest}
\caption{A taxonomy of the LLM-GGA domain with representative examples.}
\label{fig:taxonomy_of_GGA}
\end{figure*}
We comprehensively analyze these six research directions of LLM-GGA to provide valuable definitions and highlight methodologies. 
{For each direction, we summarize the key insights and limitations of relevant methods, discuss their common and different properties, and conduct an in-depth analysis of remaining challenges and promising future directions.} To further explore the capabilities of LLMs reliably, this paper uses the prompting method to test the effectiveness of LLMs in tasks such as graph structure understanding, graph learning, and graph-formed reasoning. Details of the prompts and results obtained during testing are also provided. Additionally, we refine and compile commonly used and effective prompts for graph-related tasks, assisting researchers in conducting experiments. Furthermore, this paper also organizes and introduces the code for existing popular methods, benchmarks for LLM-GGA tasks, and evaluations measuring LLM performance in graph tasks to facilitate future research.

\stitle {Comparison with other surveys.} We study state-of-the-art surveys with regard to the topics of LLM+KG, Graph-RAG, and graph foundation models \cite{li2024survey}\cite{jin2024large}\cite{liu2025graph}\cite{ren2024survey}\cite{li2024graph}\cite{peng2024graph}\cite{han2024retrieval}\cite{zhang2025survey}\cite{lavrinovics2025knowledge}. We conduct a detailed comparison of these relevant surveys in Table~\ref{summary-survey}.
In addition, we provide a novel taxonomy in Figure~\ref{fig:taxonomy_of_GGA} to give a clear organization of LLM-based graph analytics methods.
Comparing the existing survey~\cite{li2024survey}, we have conducted a more comprehensive review of the latest research works in graph learning and graph representation, e.g., ~\cite{chen2024llaga}\cite{zhang2024graphtranslator}\cite{huang2024can2}\cite{fang2024gaugllm}\cite{han2024pive}\cite{xiong2024large}\cite{zhang2024extract}\cite{chen2024exploring}\cite{shang2024path}. Moreover, we make a novel contribution to summarize graph structure understanding, graph-formed reasoning, and graph-LLM-based applications, which make this survey more thorough and up-to-date. These topics are useful for exploring LLMs in graph theory and topology (e.g., complex P and NP problems), a more reliable graph-formed LLM reasoning process (e.g., tree of thought and graph of thought), and practical applications of recommendation systems and task planning in the open world.


\stitle{Our contributions and the identified challenges for future research.} In this paper, we provide a comprehensive survey of the state-of-the-art work on LLMs applied to graph data. 
We begin by delineating six critical directions in the field of LLM-GGA: graph structure understanding, knowledge graphs and LLMs, graph learning, graph-formed reasoning, graph representation, and graph-LLM-based applications. This categorization clarifies the current work and offers a guideline for future research endeavors. In each direction, we propose a structured introduction and summarization using vivid examples and offer suitable specific pipelines. We analyze the advantages and limitations of current methodologies and suggest avenues for future research. Furthermore, we organize resources related to benchmarks, evaluations, and code links within the LLM-GGA domain to facilitate further investigation by researchers. Lastly, we identify the fundamental challenges in the LLM-GGA field, which are the primary obstacles to advancing LLM in solving graph tasks, including the fundamental issue of how sequential LLM handles structural graph data, the efficiency issue of large-scale graph data, and the NP-hard problems of complex graph analytics. This clarification guides the research direction for future work on LLM-GGA.

\stitle{Roadmaps}. The organization of this paper is as follows. We first present the preliminaries and summarize the graph description language, which converts graphs into sequences before inputting them into LLMs in Section~\ref{sec.pre}. Then, we introduce six tasks of LLM-based graph analytics one by one. We present the graph structure understanding direction in Section~\ref{structure section}, {knowledge graphs and LLMs in Section~\ref{KG section}, graph learning direction in Section~\ref{learning section}, graph-formed reasoning in Section \ref{reasoning section}, graph representation in Section~\ref{representation section} and graph-LLM-based applications in Section~\ref{application section}.} In the above six directions, we clarify the tasks that LLMs can perform, discuss the methodologies, conduct a comparative analysis, and propose guidelines and principles in this direction. Following this, Section~\ref{benchmark and evaluation} introduces the popular datasets and new datasets for solving the above tasks and also provides metrics for evaluating LLMs or tasks in different directions. In Section~\ref{future direction}, we identify and discuss the current and upcoming challenges that LLM-GGA faces and promising future directions. Finally, our conclusions are presented in Section~\ref{conclusions}.

\section{Priliminary}
\label{sec.pre}
In this section, we first introduce graphs and GNNs as a paradigm of graph-based learning models. We then distinguish LLMs from previous {PLMs (Pre-trained Language Models)}, particularly from their language modeling and training strategies. Lastly, we explore graph description languages, which can convert graphs into sequential data as input to LLMs.
\subsection{ Graphs} 
Graph data represents complex relationships through nodes and edges, where nodes represent entities and edges represent their interconnections. This structure excels at modeling intricate networks such as social, biological, and transportation systems. It enables analyses like community detection\cite{fang2020survey} and shortest paths search \cite{eppstein1998finding}, offering critical insights into the dynamics of various systems. Formally, a general graph can be represented as $\mathcal{G}=(\mathcal{V},\mathcal{E})$, where $\mathcal{V}$ and $\mathcal{E}$ denote the set of nodes and edges. $\mathcal{V}=\{v_1,...,v_n\}$ where the number of nodes is $|\mathcal{V}|=n$. $\mathcal{E}=\{e_{ij}\}$ where the number of edges is $|\mathcal{E}|$ and $e_{ij}$ is an edge from $v_i$ to $v_j$. Particularly, a \underline{t}ext-\underline{a}ttributed \underline{g}raph (TAG) can be represented as $\mathcal{G}=(\mathcal{V},\mathcal{E},\mathcal{X}_v)$, where $\mathcal{V}$ and $\mathcal{E}$ denote the set of nodes and edges. $\mathcal{V}$ is paired with raw text attributes $\mathcal{X}=\{\mathcal{X}_{v_1},...,\mathcal{X}_{v_n}\}$. 
TAGs widely exist in many real applications, such as academic citation networks with paper abstracts, social media networks with user bios/posts, and e-commerce product networks with descriptions and reviews. 
{A \underline{k}nowledge \underline{g}raph (KG) can be denoted as $\mathcal{G} = \{\mathcal{V}, \mathcal{R}, \mathcal{E}\}$ with entities $\mathcal{V}$, relations $\mathcal{R}$, and edges $\mathcal{E}$. An edge is denoted as a triple and is of the form $(s, q, o)$ where $s$ is the subject, $q$ the query relation, and $o$ the object.}


\subsection{Graph Neural Networks}
Graph Neural Networks (GNNs) \cite{kipf2016semi}\cite{velickovic2017graph}\cite{wu2020comprehensive} are a type of deep learning model that can handle graph-structured data. The goal of these GNNs is to learn representations for each node, which are computed based on the node's own features, the features of the edges connected to it, the representations of its neighbors, and the features of its neighboring nodes,
\begin{equation}
    h_v^l = AGGR(h_v^{l-1},\{h_u^l-1 :u \in N_v\} ; \theta^l)
\end{equation}

where $h_v^l$ represents the representation of node $v$ in the $l$-th layer. $AGGR$ denotes the aggregation function that aggregates the representations of neighboring nodes from the previous layer \cite{zhou2020graph}\cite{wu2022graph}. For node-level tasks, e.g., node classification, the learned representations can be used directly to accomplish specific objectives. However, for graph-level tasks, e.g., graph classification, a global representation can be obtained by pooling or integrating
representations of all nodes. 

\subsection{Large Language Models}
According to the pioneering surveys \cite{zhao2023survey}\cite{yang2024harnessing}\cite{huang2024trustllm} on LLMs, a significant distinction between LLMs and PLMs is model size. LLMs with billion-level parameters that are pre-trained on massive amounts of data, such as Llama\cite{touvron2023llama} and ChatGPT\cite{achiam2023gpt}. Conversely, PLMs are pre-trained language models with million-level parameters that can be more easily fine-tuned on task-specific data. {Additionally, the language modeling strategies and training methods are also different.
\\
\emph{\textbf{Causal Language Modeling (CLM)}} is the mainstream learning strategy of LLMs~\cite{he2023harnessing, chen2023token}, where LLMs are trained to predict the next token $x_i$ in a sequence $x=\{x_1,x_2,...,x_q\}$ based on prefix tokens $x_{<i}=\{x_1,x_2,...,x_{i-1}\}$. CLM is commonly used to train LLMs like GPT~\cite{achiam2023gpt} and Llama~\cite{touvron2023llama}. 
LLM is typically trained to optimize a conditional probability distribution $p(x_i|x_{<i})$~\cite{he2023harnessing}, which assigns a probability to each possible $x'_i \in \mathcal{D}$ given prefix tokens $x_{<i}$, where $\mathcal{D}$ is the LLM vocabulary. 
Thus, the probability of the output sequence $x$ is:
\begin{equation}
\vspace{-0.1cm}
\label{llmprob}
    p(x) = \prod\limits_{i=1}^q p(x_i|x_{<i}).
\vspace{-0.1cm}
\end{equation}
Notably, the probability of generating token $x_i$ depends only on the prefix tokens $x_{<i}$, showing that LLMs are blind to the following tokens $x_{>i}$ after $x_i$.
\\
\emph{\textbf{Instruction tuning}} is the common training method for LLMs, which aims to fine-tune the pre-trained LLMs over datasets that consist of groups of natural language instructions paired with responses~\cite{ouyang2022training}\cite{zhang2025gpt4roi}\cite{wang2023far}. The objective of instruction tuning is usually the cross-entropy loss function $\mathcal{L}$. $x$ and $y$ denote LLM's input instructions and target responses, respectively. $\theta$ refers to the LLM parameters. The loss function is as follows:
\begin{gather}
    P_\theta (y_j | x, y_{<j} ) = LLM_\theta (x, y_{<j}), \\
    \mathcal{L}_\theta=-\sum\limits_{j=1}^{|y|}\log P_\theta(y_j|x,y_{<j}).
\end{gather}
}
\begin{figure*}
    \centering
     \includegraphics[width=\textwidth]{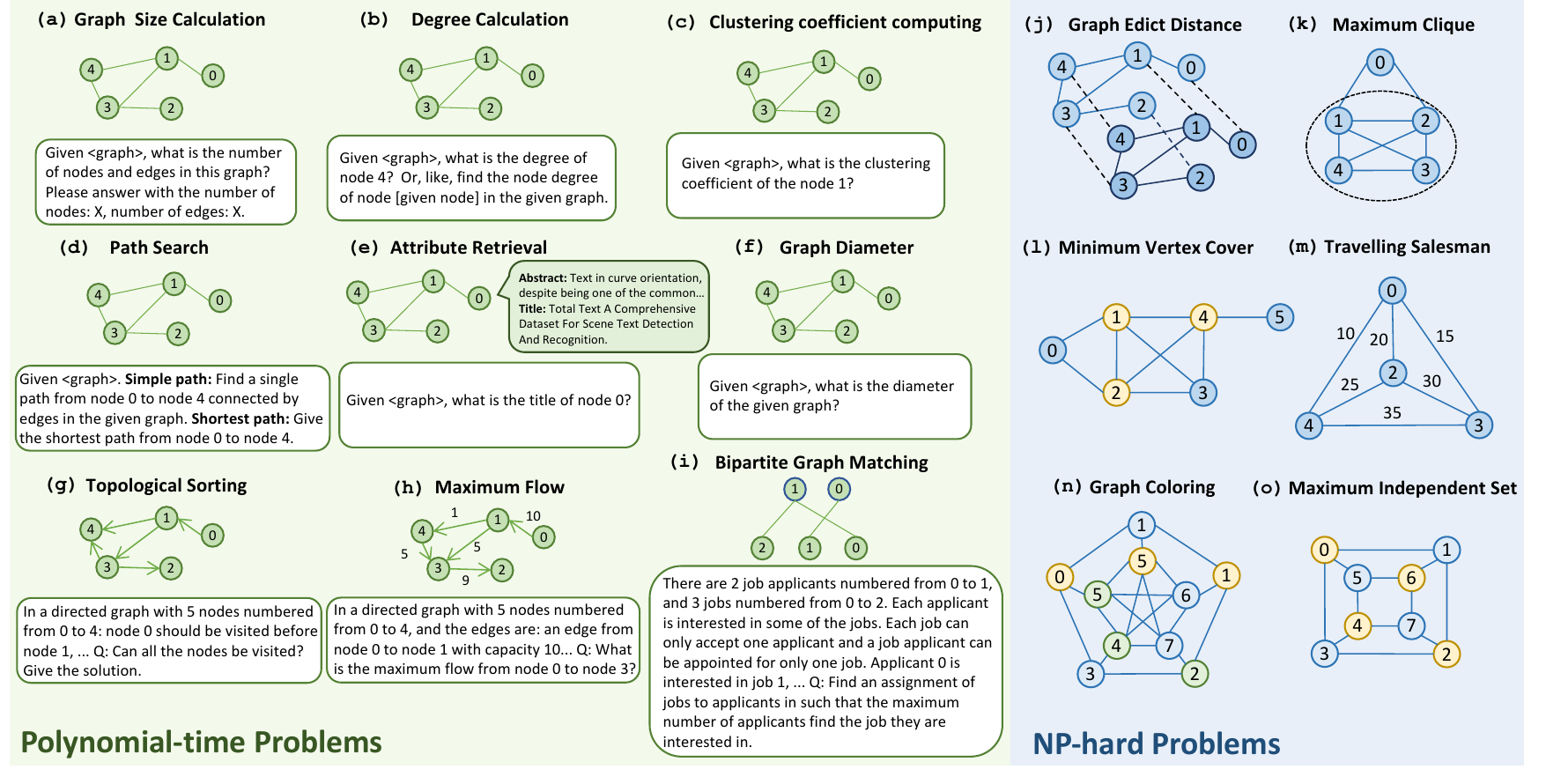} 
      \caption{Examples of graph structure understanding tasks (polynomial-time problems are shown in green and NP-hard problems are shown in blue).}
      \label{understanding-task} 
\end{figure*}
\subsection{Graph Description Language}
\label{graph description language}
Graphs are represented in structured data in arbitrary shapes, while LLMs typically process sequential data like texts. To bridge this gap, the graph description language (GDL) transforms the graph into sequential data, which can be input into LLMs. Specifically, GDL aims to convert graphs into sequential data while retaining the structure and unique attributes of the graph. This conversion allows graphs to be fed into an LLM for processing.
There are several GDLs: 
\begin{itemize}
    \item \textbf{Text description.} Graph structure can be described using words such as ‘Node 1 is connected to Node 2’ and ‘There are three nodes connected to Node 1’.
    \item \textbf{Adjacency list.} An adjacency list represents each vertex in the graph with the collection of its neighboring vertices or edges. {Node A is connected with node B and node C, which can be denoted as $\mathcal{N}(A)=\{B, C\}$.} 
    \item \textbf{SQL.} Several specialized SQL languages are designed for working with graph data, which are also capable of serving as GDLs. Some notable examples include Cypher\cite{francis2018cypher}, a query language developed by Neo4j, and Gremlin\cite{rodriguez2015gremlin}, SPARQL\cite{perez2009semantics}, and GSQL\cite{wang2019rat}. 
\end{itemize}

{There are other GDLs, such as edge lists, GML, GraphML, multi-modality encoding, and encoding as a story~\cite{shang2024survey}.}
\\
\textbf{Discussions.} {(1) Different GDLs can lead to varying results with LLMs, so testing multiple GDLs and selecting the best-performing one is recommended. (2) Adjacency lists can describe larger graphs with dozens of nodes, while other GDLs usually describe smaller ones with several nodes. (3) Natural language descriptions, like the text description and encoding as a story, can effectively make LLMs understand graph structures. Structured descriptions (e.g., SQL and GML) can more precisely represent graph structures but pose greater challenges for LLMs to comprehend. (4) API call prompts in Prompt III-6 can potentially handle large-scale graphs by calling relevant APIs. (5) Specifying the LLM's output format in the prompt can help reduce unnecessary reasoning processes.}
\section{Graph structure understanding tasks}
\label{structure section}

Graph structure understanding tasks evaluate whether LLMs can comprehend graph structures consisting of polynomial-time problems and NP-hard problems, as shown in Figure \ref{understanding-task}. Polynomial-time problems involve analyzing neighbors, shortest paths, {graph diameter, the clustering coefficient, and topological sorting}. On the other hand, NP-hard problems entail addressing tasks such as graph edit distance, maximum clique, maximum vertex cover, and other related problems. {The main challenges are: 1) comprehending graph structures locally and globally for LLMs is challenging, as LLMs are sequential models; 2) tasks with problematic assumptions and constraints are complex for LLMs to solve; 3) LLMs require algorithmic skills or strong reasoning abilities to obtain optimal or approximate answers}
This section provides an overview of various graph understanding tasks, along with their respective definitions. Furthermore, we also explore the prompts for LLMs to solve graph understanding tasks (shown in Table~\ref{prompts-for-understanding}), as well as their generalization capabilities on graph data.

\begin{table}[h!t]
\centering
\renewcommand\arraystretch{1.2}
\caption{Prompts for Graph Understanding Tasks.}
\begin{tabular}{c|l|l}
\hline
Task Categories &  Reference &  Representative Tasks  \\
\hline
\multirow{5}*{P Problems} & Prompt III-1  & graph degree   \\
\cline{2-3}
~ & Prompt III-2 & shortest path  \\
\cline{2-3}
~ & Prompt III-3 & dynamic graphs  \\
\cline{2-3}
~ & Prompt III-5 & clustering coefficient computing  \\
\cline{2-3}
~ & Prompt III-6 & graph diameter  \\
\hline
NP-hard problems &  Prompt III-4  & graph coloring    \\
\hline
\end{tabular}
\label{prompts-for-understanding}
\end{table}
\begin{figure*}
    \centering
     \includegraphics[width=\textwidth]{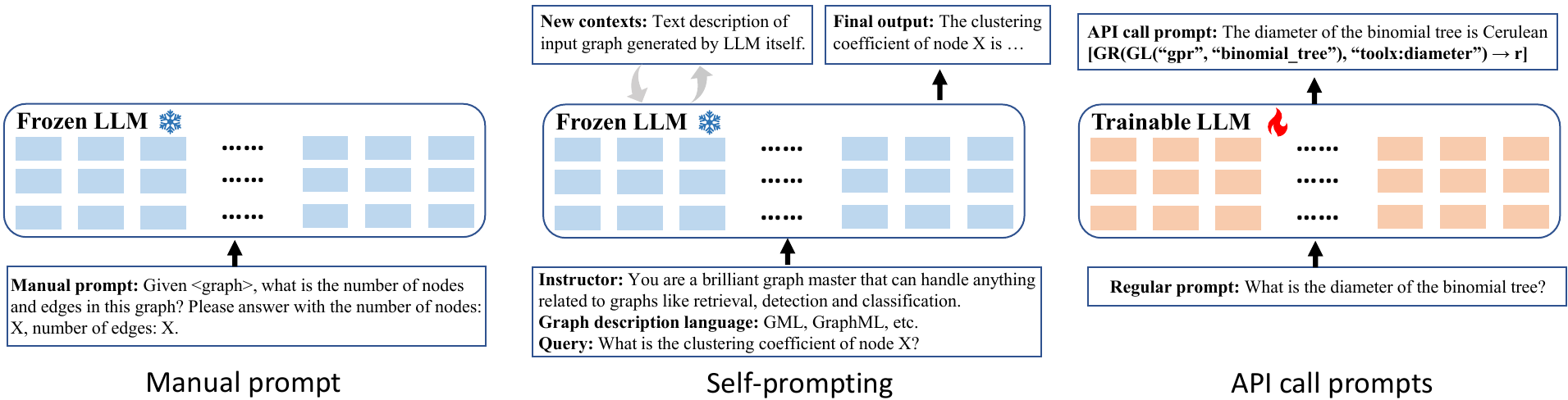} 
      \caption{Three main prompting methods in graph structure understanding tasks, which are manual prompts, self-prompting, and API call prompts.
      }
      \label{strucutre-prompting} 
\end{figure*}
\subsection{Task Introduction}
\subsubsection{{\textbf{Polynomial-time problems}}}
A problem is classified as a polynomial‐time problem if an algorithm can solve an instance of the problem in time bounded by a polynomial function of the input size. If the input size is denoted as $n$, then an algorithm runs in polynomial time if it completes in $O(n^k)$ time for some constant $k$. For example: 

\stitle{Clustering coefficient computing.} The clustering coefficient is a measure of how connected a vertex's neighbors are to one another, as shown in Figure \ref{understanding-task} (c). 
LLMs can calculate the clustering coefficient as a measure of the degree to which nodes in a graph tend to cluster together.

\stitle{Path search.} 
Given a weighted directed acyclic graph $\mathcal{G}=\{\mathcal{V},\mathcal{E}\}$ with each edge $e\in\mathcal{E}$ has a non-negative weight $w(e)$, the shortest paths task involve finding a path $p=(e_1,e_2,…,e_n)$ from a source node to a target node in $\mathcal{G}$ such that the sum of the weights of edges $w(p)=\sum^n_{i=1}w(e_i)$ is minimized. LLMs can calculate the shortest path length and identify the qualified paths, as shown in Figure \ref{understanding-task} (d). 

\stitle{Topological sorting.} Topological sorting of a directed graph $\mathcal{G}=\{\mathcal{V},\mathcal{E}\}$ refers to a linear ordering of its nodes, where each node comes before all the nodes it points to, for example, there exists a directed edge $e_{ij}$ from $v_i$ to $v_j$, $v_i$ comes before $v_j$ in the ordering. The resulting array of node ordering is called topological ordering. LLM is required to generate a valid topological sorting for the given directed graph, and there may be multiple valid solutions, as shown in Figure \ref{understanding-task} (g).

Figure~\ref{understanding-task} illustrates various tasks, including computing graph properties (size, degree, density, eccentricity, radius, diameter, and periphery), retrieval tasks (searching for connected nodes, attributes, and cycles), and more complex tasks (topological sorting, maximum flow, and bipartite graph matching). 

\subsubsection{{\textbf{NP-hard problems}}}
NP-hard problems are defined as those computational challenges for which no polynomial-time solution exists, contingent upon the assumption that $P \neq NP$.
Various NP-hard tasks like graph edit distance, graph coloring, maximum cliques, and others are shown in Figure~\ref{understanding-task}~\cite{fan2023nphardeval}\cite{tang2025grapharena}.

\stitle{Graph edit distance.} For two graphs $G_1$ and $G_2$, determine the minimum edit distance via node mappings, as shown in Figure~\ref{understanding-task} (j). The task involves aligning nodes in $G_1$ and $G_2$, and minimizing the edit operations required. The operation includes adding, deleting, or substituting a single edge or node.

\stitle{Graph coloring.} Given a graph $G = \{V, E\}$, assign colors to $V$ so that no adjacent vertices share the same color, as shown in Figure~\ref{understanding-task} (n). The objective is to minimize the number of colors used, known as the graph’s chromatic number.

\stitle{Maximum clique.} Given a graph $G = \{V, E\}$, identify the largest clique, as shown in Figure~\ref{understanding-task} (k). The task entails (1) finding a feasible clique $C \subseteq V$, and (2) ensuring that $C$ is the largest among all possible cliques.

\subsection{Graph Structure Understanding Methods}
Existing efforts have introduced various benchmarks to evaluate LLM graph reasoning potential, aiming to explore their capacity to address graph structure understanding problems. Prompting methods have emerged as the primary approach to assess LLM understanding of graph structures, with some studies also focusing on fine-tuning LLMs to enhance their graph understanding abilities. Thus, two main methods are introduced: \emph{prompting} and \emph{supervised fine-tuning LLMs}.

\subsubsection{\textbf{Prompting method}}
The prompting method\cite{liu2023pre} can be categorized into three main types: manual prompt, self-prompting, and API call prompt, as shown in Figure~\ref{strucutre-prompting}. Most studies utilize manual prompts, where carefully crafted prompts guide LLMs to comprehend graph structures better and understand the objectives of graph tasks, thereby leading to improved performance on graph-related tasks.

\textbf{Manual prompts.} NLGraph\cite{wang2024can} introduces a benchmark aiming to assess the understanding capabilities of LLMs in processing textual descriptions of graphs, covering various tasks like shortest path and maximum flow. Additionally, two prompt methods are proposed
as shown below.
\begin{framed}
\flushleft{\textbf{Prompt III-1: Build-a-Graph Prompting}} \\
Given \textless graph description\textgreater. \textbf{\emph{Let's 
construct a graph with the nodes and edges first.}} Q: What is the degree of node 4?
\end{framed}
\begin{framed}
\flushleft{\textbf{Prompt III-2: Algorithmic Prompting}} \\
\textbf{\emph{We can use a Depth-First Search (DFS) algorithm to find the shortest path between two given nodes in an undirected graph. $<$Description of DFS algorithm$>$}} \\
Given \textless graph description\textgreater. Q: Give the shortest path from node 0 to node 4.
\end{framed}
NLGraph\cite{wang2024can} shows that LLMs indeed possess preliminary graph structure understanding abilities. 
However, LLMs perform poorly on graph structures such as chains and cliques. 
To explore whether LLMs can truly comprehend graph structures and reason on graphs, {Huang et al.\cite{huang2024can} and Liu et al.\cite{liu2023evaluating} evaluate LLMs using manual prompts. Huang et al.\cite{huang2024can} examine two potential factors affecting performance: data leakage and homogeneity.} 
{Liu et al.\cite{liu2023evaluating} introduce new evaluation metrics, comprehension, correctness, fidelity, and rectification, to assess LLM proficiency in understanding graph structures.}
Beyond static graphs, LLMs' ability to understand dynamic graph structures is also assessed. Dynamic graphs change over time, capturing temporal network evolution patterns. LLM4DyG \cite{zhang2024llm4dyg} introduces a benchmark, which uses prompting methods to evaluate LLM spatio-temporal understanding capabilities on dynamic graphs, 
as shown below:
\begin{framed}
\flushleft{\textbf{Prompt III-3: DST2}} \\
\textbf{\emph{DyG Instruction:}} 
In an undirected dynamic graph, (u, v, t) means that node u and node v are linked with an undirected edge at time t.\\
\textbf{\emph{Task Instruction:}} 
Your task is to answer when two nodes are first connected in the dynamic graph. Two nodes are connected if there exists a path between them.\\
\textbf{\emph{Answer Instruction:}} Give the answer as an integer number at the last of your response after ’Answer:’\\
\textbf{\emph{Exemplar:}} Here is an example: Question: Given an undirected dynamic graph with the edges [(0, 1, 0), (1, 2, 1), (0, 2, 2)]. When are node 0 and node 2 first connected? Answer:1   \\
\textbf{\emph{Question:}}Given an undirected dynamic graph with the edges [(0, 9, 0), (1, 9, 0), ...]
When are node 2 and node 1 first connected?
\end{framed}
Results show that LLMs have preliminary spatio-temporal understanding capabilities on dynamic graphs. Dynamic graph tasks become increasingly challenging with larger graph sizes and densities, while being insensitive to periods and data generation mechanisms.
Investigations into LLM's performance extend P problems to NP-hard problems. {Dong et al.\cite{dong2023large} delve into the potential of LLMs to address the classical P vs. NP question.} Furthermore, NPHardEval~\cite{fan2023nphardeval} and GraphArena~\cite{tang2025grapharena} propose benchmarks for P and NP-hard problems, such as graph edit distance, graph coloring and others. For instance, the prompt for the graph coloring problem is shown below:
\begin{framed}
\flushleft{\textbf{Prompt III-4: Graph coloring. (NP-hard problem)}}\\
\textbf{\emph{Task Instruction:}} $\langle$Graph coloring description$\rangle$. There are 6 vertices 1 to 6 in a graph. You may use 4 colors with alphabets from A, B, C,...to color the graph. Please label every vertex, even if it is disconnected from the rest of the graph. Please provide each vertex's color.\\
\textbf{\emph{Answer Instruction:}} Your output should contain two parts enclosed by $<$root$><$/root$>$...\\ 
\textbf{\emph{Given Graph:}} Vertex 1 is connected to vertex 6,...
\end{framed}
The work mentioned above provides an initial exploration of the cognitive capabilities of LLMs regarding various graph structures, including their performance on P and NP-hard problems. 
Some other studies conduct in-depth investigations into the generalization capabilities of LLMs, examining whether they rely on reasoning or memory to address questions related to graph structures. NLGIFT~\cite{zhang2024can} proposes a benchmark to assess the generalization capabilities of LLMs, explicitly investigating whether LLMs can surpass the semantic, numeric, structural, and reasoning patterns present in the graph data.
To explore LLM sensitivity to different graph-related prompts, GraphInsight~\cite{cao2024graphinsight} posits that LLMs exhibit a ``position bias" in their understanding of graphs when prompted. 

\textbf{Self-prompting.}
{The self-prompting method refers to the process where an LLM designs prompts for itself based on the original prompt. Specifically, the LLM continuously updates the initial prompt to make it easier for itself to understand contexts in prompts (e.g., GSQL, Cypher) and the goal of the task.} GPT4Graph\cite{guo2023gpt4graph} utilizes self-prompting by continuously guiding LLMs to refine the prompt with descriptions of graphs, converting the original structured GDLs (i.e., GSQL) into LLM-generated text descriptions (as shown in Section \ref{graph description language}). 
\begin{framed}
{\flushleft{\textbf{Prompt III-5: Self-prompting}} \\
\textbf{\emph{Instructor:}}
You are a brilliant graph master that can handle anything related to graphs like retrieval, detection and classification.\\
\textbf{\emph{Graph description language (Original prompt):}} GSQL, GML, GraphML~\cite{shang2024survey} that describe graph structures.\\
\textbf{\emph{Self-updated prompts:}} Node P357 has 4 neighbors, where each of which are about anomaly detection... \\
\textbf{\emph{Query:}} What is the clustering coefficient of node P357?}
\end{framed}

\begin{figure}
    \centering
     \includegraphics[width= 0.47\textwidth]{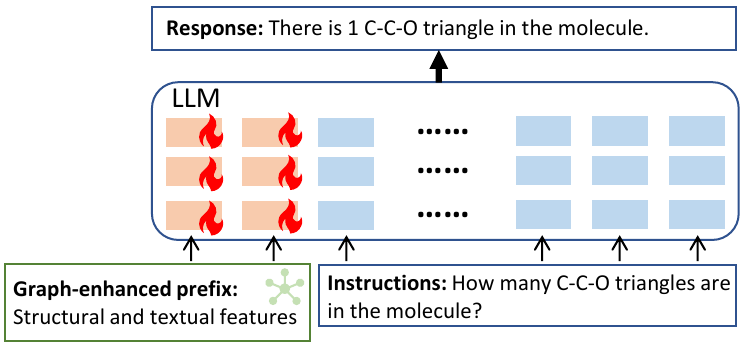} 
      \caption{Supervised fine-tuning methods for graph structure understanding tasks, like prefix tuning\cite{chai2023graphllm}. 
      }
      \label{SFT-structure} 
\end{figure}

{\textbf{API call prompts.} The API call prompts aim to address LLMs’ limited ability in topological structuring and temporal information processing by calling APIs~\cite{patil2024gorilla}\cite{zhang2023graph}, enhancing LLM-based agents. Specifically, the API call prompt is generated by LLMs to convert complex and difficult tasks into a rule-based logical expression like \emph{GL("benzenering")} in Prompt III-6. The rule-based logical expression corresponds to prepared API calls. Inspired by recent models such as Toolformer\cite{schick2024toolformer}, Graph-ToolFormer\cite{zhang2023graph} is proposed to equip LLMs such as GPT-J\cite{wang2021gpt} and LLaMA\cite{touvron2023llama1}\cite{touvron2023llama} with graph structure understanding and reasoning capabilities by calling graph reasoning APIs of external tools. Gorilla\cite{patil2024gorilla} employs a fine-tuned LLaMA enhanced with Retriever Aware Training (RAT), surpassing GPT-4 in writing API calls and significantly reducing hallucinations. One API call prompt is shown below.}
\begin{framed}
{\flushleft{\textbf{Prompt III-6: API call prompts}} \\
\textbf{\emph{Input:(Regular prompt)}} \\
The structure of the benzene ring molecular graph of benzene ring contains a hexagon.\\
\textbf{\emph{Output:(API call prompt)}} \\ 
The structure of the 
\textbf{[GL("benzenering")]} molecular graph of benzene ring contains a hexagon.}\\
\end{framed}

\subsubsection{\textbf{Supervised fine-tuning (SFT) method}}
Beyond leveraging prompts for graph-structured tasks with LLMs, certain studies also explore LLM-supervised fine-tuning, illustrated in Figure \ref{SFT-structure}. GraphLLM\cite{chai2023graphllm} introduces a hybrid model that inherits the capabilities of both graph learning models and LLMs, enabling LLMs to reason on graph data proficiently while utilizing the superior power of graph learning models.

\subsection{Summary of Methods, Challenges and Future Directions}
\label{structure prompt}
{In summary, the prompting method can be divided into three categories: manual prompts, self-prompting, and API call prompts. Most current methods primarily rely on manual prompts, incorporating techniques like Chain of Thought (CoT)\cite{wei2022chain}, self-consistency\cite{wang2022self}, and in-context learning\cite{dong2024survey}. Self-prompting methods are also widely used to obtain better prompt representations. However, relying solely on manual prompts and self-prompting provides only marginal enhancements to model performance, as these approaches primarily leverage the existing capabilities of LLMs. Additionally, due to the limited input window of LLM, the graph size that can be input to LLM at once is also restricted, while graph sizes in the real world are typically large. }

{To overcome these shortcomings, we propose two feasible future directions to better leverage existing LLMs for handling structure understanding tasks. The first direction is breaking down complex tasks into several sub-problems. While LLMs can tackle simple graph tasks, they struggle with more challenging ones. Breaking down complex graph understanding tasks into simpler components enables LLMs to engage in multi-step reasoning processes, leading to the resolution of complex issues, such as GoT\cite{zhang2023graph}, which can help address more intricate graph tasks like generating GNN frameworks, k-truss tasks, kd-core tasks, etc. The second direction is API call prompts. Inspired by ToolFormer\cite{schick2024toolformer}, LLMs can be trained as agents to utilize tools for graph tasks that are hard to solve. However, current API call prompt methods\cite{zhang2023graph} utilize LLMs solely to convert user queries into API command strings for processing by subsequent programs, like \textbf{Prompt III-6}.}

Compared to prompting methods, fine-tuning LLMs with graph data is a more practical approach for enhancing their comprehension of graph structures. Current methods mainly utilize SFT to train LLMs, enabling them to grasp the entire graph structure through instruction tuning techniques. Various other training techniques in the natural language processing field, such as RLHF, RLAIF, PPO, and DPO, can also be employed to understand graph structures, aligning LLMs with human preferences. RLHF\cite{ouyang2022training} provides detailed human feedback through pairwise comparison labeling. Furthermore, to address the instability issue in PPO\cite{schulman2017proximal} training, RAFT\cite{dong2023raft} can also be attempted, which requires online interaction. For offline algorithms, methods like DPO\cite{rafailov2024direct} and PRO\cite{song2024preference} can also be utilized for training LLMs.
\\
\textbf{Discussions.} 
We further discuss the cognitive and generalization capabilities of LLMs on graphs. {Regarding cognitive abilities, LLMs can handle simple P and NP-hard problems, commonly with several nodes. When solving P or NP problems, LLMs can often provide direct answers. For more complex problems, they use common algorithms and step-by-step reasoning, such as BFS or DFS for shortest path problems. Particularly, exhaustive enumeration is a common strategy chosen by LLMs. Moreover, their performance is not a result of data leakage in the pre-training corpus.} However, they lag behind traditional graph algorithms in both efficiency and accuracy. 
{Regarding generalization capabilities, current evidence indicates that although LLMs exhibit strong robustness and generalization performance in handling variations in semantic and numerical attributes of graphs and various graph analytical tasks, they still face challenges in large-scale and complex networks.}

\section{{Knowledge graphs and LLMs}}
\label{KG section}
\begin{figure}
    \centering
     \includegraphics[width= 0.46\textwidth]{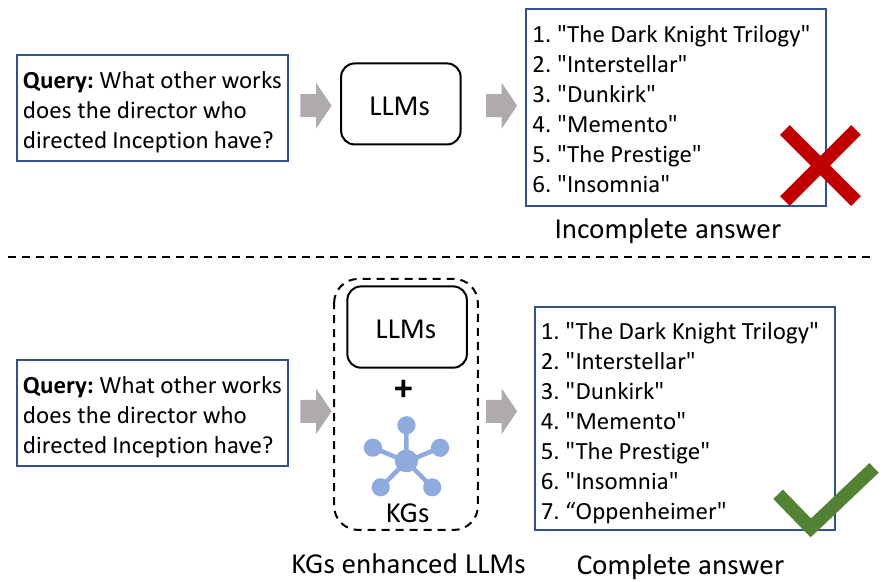} 
      \caption{Knowledge graphs can enhance LLMs to provide more comprehensive answers to tackle hallucinations.}
      \label{KG-LLM} 
\end{figure}

LLMs have shown remarkable reasoning capabilities in challenging tasks, sparking debates on the potential replacement of Knowledge Graphs (KGs) in triplet form (subject, predicate, object) by LLMs. Recent LLMs are seen as viable alternatives to structured knowledge repositories such as KGs, indicating a shift towards utilizing LLMs for processing real-world factual knowledge \cite{sun2024head}\cite{yang2024give}. {However, 
LLMs, against KGs, face several significant challenges: 1) Hallucination is a common issue for LLMs due to a lack of domain-specific knowledge and knowledge obsolescence, leading to incorrect reasoning and reduced credibility in critical scenarios like medical diagnosis and legal judgments \cite{sun2024head} \cite{pan2024unifying} \cite{luo2024reasoning}. Although some LLMs can explain predictions through causal chains, they struggle to address hallucination effectively. Integrating external KGs can help mitigate these problems\cite{sun2024think}. 2) Insufficient domain knowledge hampers LLM performance in specific areas, including private datasets, necessitating the integration of domain-specific knowledge graphs to enhance their ability to answer domain-specific questions \cite{dernbach2024glam}. 3) LLMs struggle with recalling facts when generating knowledge-based content, despite excelling in learning language patterns and conversing with humans\cite{yang2024give}. 4) LLMs have limitations in accurately capturing and retrieving knowledge, hindering their ability to access factual information effectively\cite{tian2024graph}.}
Thus, KGs like Wikipedia and DBpedia are structured repositories of rich factual knowledge, providing a more explicit and reliable source of information, 
as shown in Figure~\ref{KG-LLM}. 

\subsection{KG Solutions to tackle LLM Limitations}
{To address the limitations of LLMs, such as hallucination, insufficient domain knowledge, etc., integrating LLMs with external KGs is a potential way to allow LLMs to reason on high-quality knowledge, thereby enhancing their capabilities. Retrieval augmented generation (RAG) based on KG is currently the mainstream technique of KG solutions to tackle LLM limitations. It aims to dynamically query large-scale textual databases to retrieve relevant factual knowledge and integrate it into the responses generated by LLMs.}
Existing RAG works focus on addressing the above four limitations of LLMs, which are \emph{hallucination}, \emph{insufficient domain knowledge}, \emph{struggling with recalling facts}, and \emph{roughly capturing knowledge}.

\stitle{Hallucination.} To address the hallucination issues in LLMs, {the Head to Tail benchmark\cite{sun2024head} is introduced to assess LLM reliability in answering factual questions and to evaluate the probability of hallucination in generating KG triples. 
G-retriever\cite{he2024g} performs a Prize Collecting Steiner Tree optimization problem to search for relevant knowledge from KGs with closest and minimal structures.
ToG\cite{sun2024think} partially addresses hallucination by involving the LLM agent in iteratively searching KGs, identifying promising reasoning paths, and providing likely reasoning outcomes. RoG\cite{luo2024reasoning} synergizes LLMs with KGs for faithful and interpretable reasoning.} 

\stitle{Insufficient domain knowledge.} The second limitation is that LLMs need domain-specific knowledge. To tackle this, GLaM\cite{dernbach2024glam} is developed to convert knowledge graphs into text paired with labeled questions and answers, allowing LLMs to acquire and respond to domain-specific knowledge. 

\stitle{Struggling with recalling knowledge.} Regarding the limitation related to LLMs forgetting facts, KGPLMs\cite{yang2024give} is introduced to enhance the model's ability to recall facts compared to standalone LLM, where LLMs improve knowledge extraction accuracy, and KGs guide LLM training to enhance memory and knowledge application capabilities.

\stitle{Roughly capturing knowledge.} Finally, the fourth limitation pertains to LLM challenges in accurately retrieving and returning knowledge from KGs. KGs can enhance LLM performance by incorporating them during pre-training and inference stages or to deepen LLM's understanding of acquired knowledge. {GNP\cite{tian2024graph} proposes a plug-and-play method to facilitate pre-trained LLMs in effectively learning beneficial knowledge from KGs. A graph retrieval algorithm based on text indexing or graph indexing to obtain relevant knowledge is also an effective method. GraphRAG~\cite{edge2024local} divides the KG into several communities through a community detection algorithm and then employs LLMs to generate summaries for closely related communities. ToG-2\cite{ma2024think} integrates graph retrieval and context retrieval to obtain in-depth clues relevant to the question, enabling LLMs to generate reliable answers.}

\stitle{Summary and discussion.} {Existing research on RAG faces two main limitations: multi-modality KG handling and efficient retrieval framework on large-scale KGs. Current efforts primarily focus on textual information in KGs, exhibiting limited transferability to other modalities such as images and audio. Given massive multi-modality information contained in KGs, effectively retrieving closely relevant knowledge is challenging. Moreover, in large-scale KGs, text indexing is particularly time-consuming, taking tens of times longer than the subsequent retrieval process. As existing methods concentrate on small-scale KGs with thousands of entities~\cite{peng2024graph}, efficient retrieval algorithms also become difficult as the scale of KGs grows.}

\subsection{LLM Solutions for KG Tasks}
LLMs can enhance KGs to tackle a broader array of challenges. By leveraging LLMs, KGs can be fortified to perform various KG-related tasks such as embedding, completion, construction, text generation from graphs, and question answering~\cite{pan2024unifying}. An illustrative example is how LLMs can support KG tasks such as knowledge graph alignment. In entity alignment tasks between different knowledge graphs, the objective is to identify pairs of entities representing the same entity. To address this, AutoAlign~\cite{zhang2023autoalign} 
automatically identifies similarities between predicates across different KGs with LLM assistance. 
To improve the LLM performance in question answering, {Shah et al.\cite{shah2024improving} enhance} LLMs to generate Cyber and SPARQL queries for multi-hop knowledge graph question answering. To achieve more accurate knowledge inference while minimizing the costs associated with frequent interactions between LLMs and KGs, EtD~\cite{liu2024explore}
employs GNNs to identify promising candidate answers and fine-grained knowledge relevant to the question,
then guides the frozen LLM to ascertain the final answer. For KG completion (KGC) tasks, {Sehwag et al. \cite{sehwag2024context} integrate} ontological knowledge and graph structure through in-context learning to improve KGC performance. To tackle the problem of infrequent and less popular relationships substantially hindering KGC performance, MuKDC~\cite{li2024llm} utilizes multi-level knowledge distillation and generates supplementary knowledge to mitigate data scarcity in few-shot environments. 
Furthermore, the combination of KGs and LLMs 
can address tasks like multi-document question answering~\cite{wang2024knowledge} and fact-checking in public deliberation~\cite{giarelis2024unified}. 
Overall, there exist several comprehensive surveys~\cite{pan2024unifying}\cite{zhu2024llms}\cite{yang2024give} on the study of LLMs and KGs for further investigations.
\section{Graph learning tasks}
\begin{figure*}
    \centering
     \includegraphics[width=\textwidth]{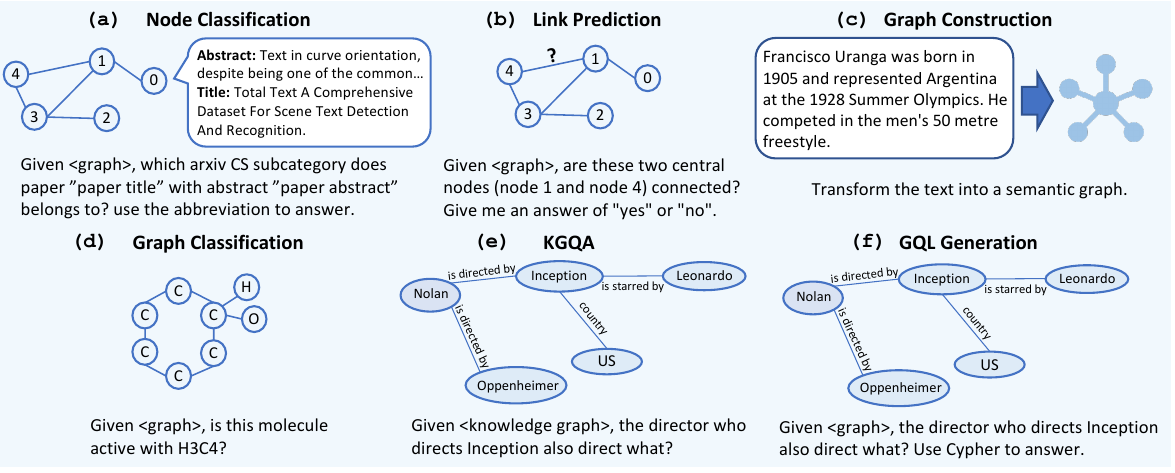} 
      \caption{Examples of graph learning tasks.}
      \label{graph-learning} 
\end{figure*}
\label{learning section}

This section introduces various graph learning tasks that LLMs can address, as shown in Figure~\ref{graph-learning}, such as node classification, graph construction, etc. {When adapting LLMs to tackle these tasks, the main challenges are: 1) nodes are numerous with complex text attributes, which are challenging for LLMs to analyze; 2) graph foundation models require handling diverse, scalable and complicated networks and structures like social network, biomedical network, citation network, etc; 3) graph foundation models require handling various multi-mode downstream tasks, like classification, reasoning, and description.} To address these challenges, we discuss technique details, key insights, and limitations of relevant and representative methods, which primarily leverage LLMs' extensive knowledge base, semantic comprehension, and generalization capabilities. {Specifically, Table~\ref{summary-learning} shows a detailed comparison of six influential works, which are TAPE\cite{he2023harnessing}, OFA\cite{liu2024one}, LLaGA\cite{chen2024llaga}, GraphGPT\cite{tang2024graphgpt}, TG-LLM\cite{xiong2024large} and PiVe \cite{han2024pive}. It directly highlights their common insights and differences from perspectives of the role of LLMs, the training of LLMs, the usage of GNNs, the downstream tasks, and the training strategy.} 

\subsection{Task Introduction}
Graph learning tasks include node classification, link prediction, graph classification, graph construction, knowledge graph question answering (KGQA), and graph query language (GQL) generation, as shown in Figure~\ref{graph-learning}. Below, we detail several prominent graph learning tasks. 

\stitle{Node classification} requires LLMs to learn from the neighbors or attributes of a node. This process involves classifying unseen nodes within a given graph. For example, in an academic network, this task might involve categorizing papers into different research directions, as depicted in Figure~\ref{graph-learning} (a).

\stitle{Link prediction} requires LLMs to determine whether two central nodes are connected, given their text attributes and neighbor nodes, as shown in Figure~\ref{graph-learning} (b).

\stitle{Graph construction.} Given a natural language paragraph, LLMs are required to extract the entities and relationships from textual data and subsequently construct a graph based on the extracted information, as shown in Figure~\ref{graph-learning} (c).

\begin{table*}[h!t]
\scriptsize
   \centering
   \renewcommand\arraystretch{1.2}
   \setlength\tabcolsep{8pt}
   \caption{Comparison of representative works in graph learning.}
   \begin{tabular}{l|c|c|c|c|c}\hline
    Method &  Role of LLMs & Training of LLMs &  GNNs usage &   Downstream tasks 
    &   Training strategy \\ \hline
    TAPE\cite{he2023harnessing} & Enhancers & \XSolidBrush &  \Checkmark & 
    Node classification 
    & Supervised\\  \hline
    OFA\cite{liu2024one} & Enhancers & \XSolidBrush  & \Checkmark & 
    Node classification, graph classification, link prediction 
    &  Supervised  \\  \hline
    LLaGA\cite{chen2024llaga} & Predictors & \Checkmark  & \XSolidBrush &
    Node classification, node description, link prediction 
    &  Supervised  \\  \hline
    GraphGPT\cite{tang2024graphgpt} & Predictors & \Checkmark  & \XSolidBrush  & 
    Node classification, link prediction 
    &  Self-supervised  \\ \hline
    TG-LLM\cite{xiong2024large} &  Generators & \Checkmark  & \XSolidBrush & 
    Temporal reasoning, temporal question-answering 
    &  Supervised\\ \hline
    PiVe \cite{han2024pive} & Generators & \XSolidBrush  & \XSolidBrush &
    Text-to-graph 
    & \XSolidBrush \\  \hline
    
    \end{tabular}
    \label{summary-learning}
\end{table*}

\begin{figure*}
    \centering
     \includegraphics[width=\textwidth]{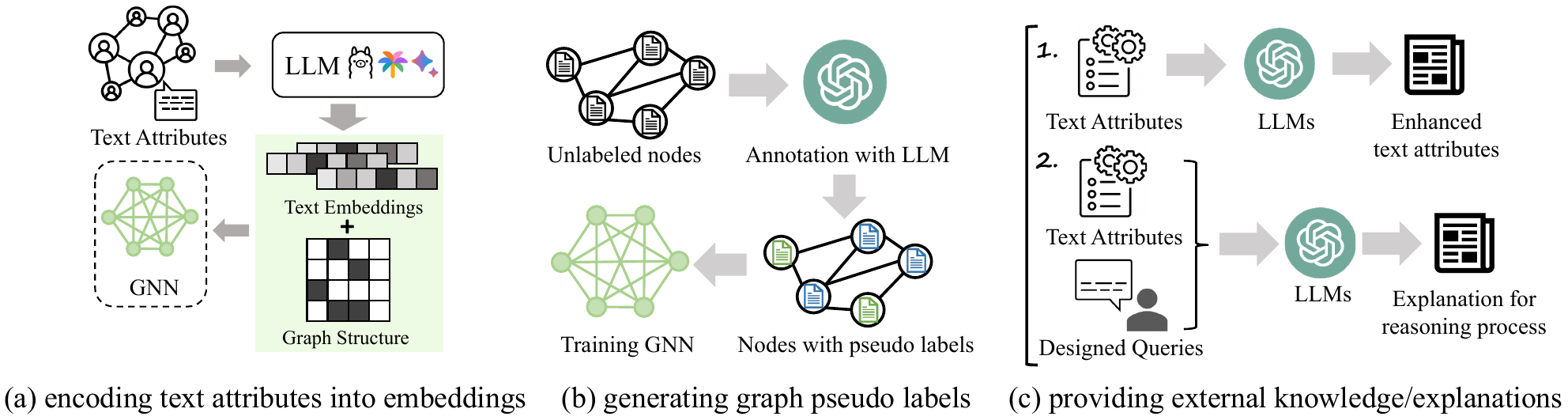} 
      \caption{LLMs-as-enhancers methods are categorized into three types: (a) Encoding text attributes into embeddings by inputting node text attributes into LLMs, then integrating these embeddings with graph structures for GNN training. (b) Generating graph pseudo-labels by feeding unlabeled nodes to LLMs for labeling, subsequently using these pseudo-labeled nodes for GNN training. (c) {Providing external knowledge for downstream GNNs through two pipelines: the first enhances text attribute details via elaboration, while the second uses text attributes and queries to generate answers and explain reasoning.}}
      \label{llm-as-enhancers} 
\end{figure*}

\subsection{Graph Learning Methods}
Concerning graph learning methods, specific techniques are designed to address individual tasks, while others can tackle various graph learning tasks. These approaches can be further categorized based on the role of LLMs in facilitating graph learning into three distinct classifications: LLMs as enhancers, LLMs as predictors, and LLMs as graph generators.
When LLMs function as enhancers, they utilize their sophisticated semantic comprehension, robust reasoning abilities, and extensive knowledge base to augment the textual attributes of nodes, thus enhancing the performance of other models, such as GNNs. Conversely, when LLMs act as predictors, they are prompted or fine-tuned to solve graph learning tasks. Furthermore, when functioning as graph generators, LLMs extract semantic graphs from natural language, thereby facilitating information extraction and knowledge graph construction.
In summary, adapting LLMs in graph learning tasks presents a promising avenue for advancing the field. By leveraging the strengths of LLMs as enhancers and predictors, researchers can explore new directions for enhanced performance and more profound insights into LLM-GIL tasks.

\subsubsection{\textbf{LLMs-as-enhancers}}
The predominant approach in the LLMs-as-enhancers is the LLM-GNN pipeline. Within this framework, LLMs are tasked with processing text attributes, while GNNs are responsible for handling graph structures, capitalizing on the complementary strengths of both components to address graph learning tasks effectively. LLMs bolster GNNs through three distinct mechanisms: encoding text attributes into embeddings (as shown in Figure~\ref{llm-as-enhancers} (a)), generating graph pseudo labels (as shown in Figure~\ref{llm-as-enhancers} (b)), and providing external knowledge or explanations (as shown in Figure~\ref{llm-as-enhancers} (c)). Subsequently, we will provide a comprehensive elaboration on these three enhancement strategies.

\textbf{Encoding text attributes into embeddings.} Many existing pipelines utilize LLMs to encode text attributes into node embeddings as node features and then feed these embeddings into a GNN for learning, as shown in Figure~\ref{llm-as-enhancers} (a). TAPE~\cite{he2023harnessing} fine-tunes LM and GNN for node classification, integrating original node text attributes and LLM explanations to derive node embeddings. GraphAdapter~\cite{huang2024can2} designs two training stages for GNN and fusion module to integrate LLM-generated embeddings for node classification. GAugLLM~\cite{fang2024gaugllm} designs expert augmented texts as new node text attributes, which are fed into LLM for deriving node embeddings and then trains GNN based on it. Moreover, SIMTEG~\cite{duan2023simteg} and GLEM~\cite{zhao2023learning} train an LM for generating node embeddings and then train a GNN based on these node embeddings for node classification. More general models can perform various graph learning tasks. OFA~\cite{liu2024one} utilizes LLMs to unify diverse graphs by describing nodes and edges in natural language and then trains a GNN based on the obtained texts for various downstream tasks.
\begin{figure*}
    \centering
     \includegraphics[width=\textwidth]{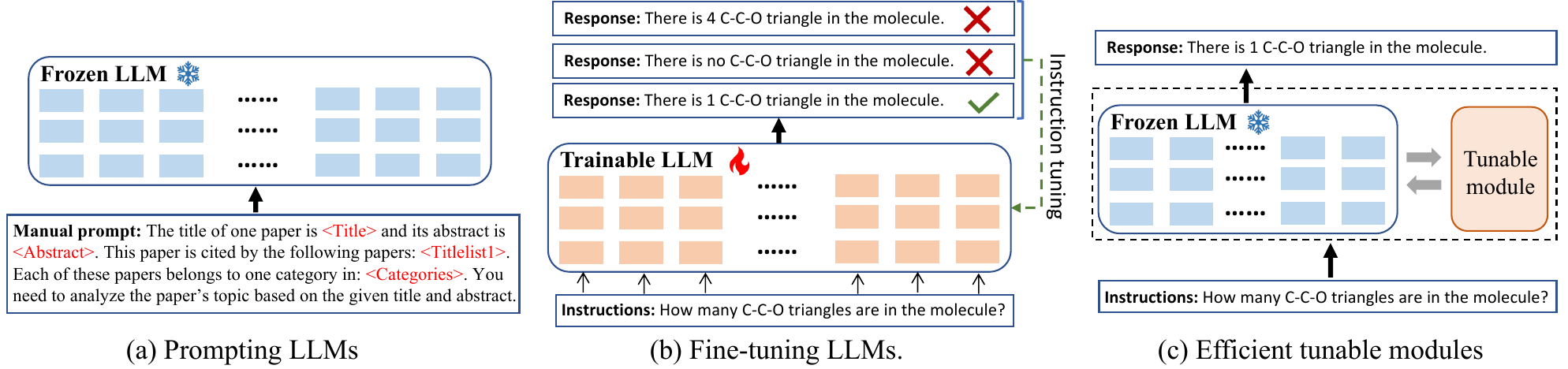} 
      \caption{LLMs-as-predictors methods are categorized into three types: (a) Prompting LLMs by inputting designed prompts to predict nodes, links, or graphs; (b) Fine-tuning by providing instructions to generate responses, followed by refining LLMs based on feedback; (c) Efficient tunable modules where frozen LLMs are combined with tunable components to tackle graph learning tasks, emphasizing the training of lightweight modules.}
      \label{predictor} 
\end{figure*}

\textbf{Generating graph pseudo labels.} 
The extensive knowledge base and powerful generalization ability of LLMs enable their utilization in generating pseudo labels for graphs for data augmentation, as shown in Figure~\ref{llm-as-enhancers} (b). This process significantly enhances the GNNs training process and improves performance in graph learning tasks. GLEM~\cite{zhao2023learning} proposes training the GNN and LM separately within a variational EM framework, where the LM predicts gold and pseudo-labels from the GNN label results.
Moreover, LLM-GNN~\cite{chen2024label} proposes to select a candidate node set for annotation by LLMs to facilitate GNN training, addressing challenges of high annotation costs and the need for large amounts of high-quality labeled data.

\textbf{Providing external knowledge/explanations.} 
{Owing to the extensive and diverse corpus that LLMs have been trained on, coupled with their robust reasoning capabilities, LLMs can provide external knowledge or explanations relevant to certain nodes to enhance the attributes of nodes, as shown in Figure~\ref{llm-as-enhancers} (c). 
The enhanced node attributes assist the downstream GNNs in better extracting and capturing node features.} Graph-LLM~\cite{chen2024exploring} utilizes LLMs like ChatGPT to generate augmented text attributes and pseudo labels for trainable LLM and GNN training.
Similarly, TAPE~\cite{he2023harnessing} utilizes LLMs to predict node labels and explain reasoning processes, which are then used as text attributes to train the LM and GNN.

\subsubsection{\textbf{LLMs-as-predictors.}}
When LLMs are predictors, they are typically employed as standalone predictors. The critical aspect of integrating LLMs as predictors lies in crafting well-designed prompts and fine-tuning methods. Well-designed prompts encompass text attributes and graph structures, enabling LLMs to comprehend graphs and improve prediction accuracy. In contrast, fine-tuning methods, such as instruction tuning, enhance the LLM's grasp of the graph structure. Thus, LLMs-as-predictors methods can be categorized into prompting LLMs and fine-tuning LLMs, {as illustrated in Figure~\ref{predictor}.}

\textbf{Prompting LLMs.} 
The prompting method can be divided into two categories. One type is the manual prompts, which are manually written prompts. For instance, Beyond Text~\cite{hu2023beyond} and KG-LLM~\cite{shu2024knowledge} utilize manual prompt templates with slots. By filling these slots with different examples, various prompts can be constructed. 
Compared to manual prompts, LPNL~\cite{bi2024lpnl} generates prompts through a two-stage sampling process. 

\textbf{Fine-tuning LLMs.} IntructGLM~\cite{ye2024language} and GraphGPT~\cite{tang2024graphgpt} fine-tune LLM by instruction tuning for node classification. 
InstructGLM~\cite{ye2024language} uses prompts to input subgraph structures into LLMs, which are then tasked with answering questions and predicting node labels. Conversely, GraphGPT~\cite{tang2024graphgpt} inputs subgraph structures into LLMs through embedding, followed by two rounds of instruction tuning for node classification.

\textbf{Efficient tunable modules.} Training LLMs is time-consuming and costly, and inappropriate training can undermine their original reasoning and generalization capabilities. Therefore, existing research attempts to design efficient tunable modules in conjunction with LLMs, focusing on training a tunable module with fewer parameters instead of the LLM itself, thereby reducing both the time and cost.
LLaGA~\cite{chen2024llaga} designs a tunable graph projector to align different graphs, which can handle various graph tasks across multiple datasets, eliminating the need for task-specific adjustments. {GraphTranslator~\cite{zhang2024graphtranslator} designs} a tunable module, named Translator, to bridge the frozen graph model and LLMs. 
ENGINE~\cite{zhu2024efficient} proposes tunable G-Ladders combined with frozen LLM layers via a side structure, eliminating the backpropagation of LLMs.

\begin{figure}
    \centering
     \includegraphics[width=0.36\textwidth]{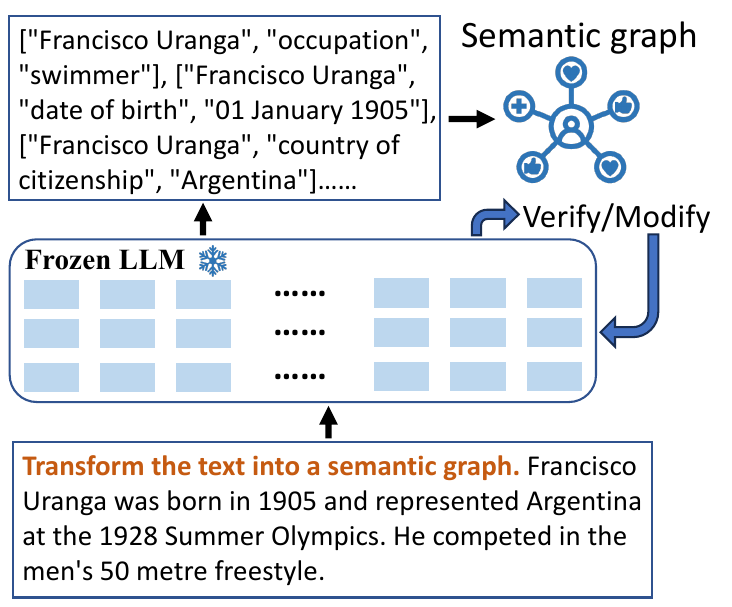} 
      \caption{LLMs-as-graph-generators: LLMs generate graphs by extracting entities and key relationships.}
      \label{generator} 
\end{figure}

\subsubsection{\textbf{LLMs-as-graph-generators}}
{Many studies explore the capability of LLMs to generate semantic graphs from natural language, thereby benefiting various applications such as decision-making and recommendation, as shown in Figure~\ref{generator}.}
PiVe~\cite{han2024pive} proposes an iterative verification framework to improve the graph generative capability of LLMs. TG-LLM~\cite{xiong2024large} fine-tunes LLMs to construct temporal graphs from text contexts and reason on temporal graphs rather than reason on text contexts. To enhance the quality of semantic graph generation from text in practical applications using LLMs, EDC~\cite{zhang2024extract} proposes a three-phase framework. This framework first extracts triples from the text and subsequently automatically modifies any inappropriate keywords within those triples.
\subsection{Summary of Methods, Challenges and Future Directions}
{In summary, for addressing graph learning tasks, existing methods~\cite{he2023harnessing}\cite{tang2024graphgpt} categorize based on the role of LLM into three types: LLMs-as-enhancers (LLM-GNN pipelines), LLMs-as-predictors (LLM pipelines), and LLMs-as-graph-generators (text-to-graph pipelines). 
\textbf{When LLMs function as enhancers}, the most popular pipeline is the LLM-GNN pipeline. There are three categories of LLM-GNN pipelines, depending on how LLM enhances GNN: encoding text attributes into embeddings, generating graph pseudo labels, and providing external knowledge/explanations. However, the LLM-GNN pipelines that are currently available are not end-to-end pipelines, meaning that LLM and GNN cannot be trained together. LLM and GNN can be trained separately using frameworks like EM framework~\cite{zhao2023learning} or by freezing LLM and using it as an external knowledge base. Co-training LLM and GNN can lead to issues like gradient vanishing, which is a significant obstacle in current LLM-GNN pipelines due to the large number of parameters in LLM compared to GNN. 
To solve these limitations, we propose feasible future directions like knowledge distillation, which can reduce the number of LLM parameters while retaining the beneficial capabilities for downstream tasks.
\textbf{When LLMs function as predictors}, three main methods are used: prompting LLMs, SFT LLMs and efficient tunable module. All approaches for fine-tuning LLMs can be reviewed in the ``comparisons and discussions" section of Section~\ref{structure section}. Currently, SFT and DPO are popular methods for fine-tuning LLMs. 
\textbf{When LLMs function as graph generators}, existing studies mainly use prompt engineering methods to iteratively prompt LLMs to extract entity-relationship triples in paragraphs. PiVe~\cite{han2024pive} and EDC~\cite{zhang2024extract} provide verifying or modifying ways to improve semantic graph generation quality.}

\stitle{Challenges and future directions.} {Classical graph tasks, such as node classification on attributed static networks, have recently obtained the most attention. However, there is potential for more complex tasks in the future, such as predicting graph evolution on dynamic graphs. Leveraging LLM models that are suitable for handling sequential data and can process time series data, along with GNNs capturing changes in graph structures, can help address a broader range of problems effectively.}
\section{Graph-formed reasoning}
\label{reasoning section}

Graph-formed reasoning refers to the process that LLMs engage in cognitive operations utilizing graph structures, such as the graph of thoughts, or verify conclusions through these structures to achieve more accurate and reliable answers. Simultaneously, this methodology aids in the interpretability analysis of the LLM reasoning process. LLMs have strong reasoning capabilities, and many prompting methods are proposed to enhance LLM reasoning abilities, addressing algorithmic problems, mathematical issues, etc., such as chain of thought, self-consistency, in-context learning, and more. {However, these methods diverge from the patterns of human thought. The human thought process typically consists of non-linear continuous thoughts. The main challenge is to design the LLM reasoning process that aligns with the human thought process.} Graphs can represent the thinking patterns of humans during the thought process. In this section, we first introduce the tasks that LLMs can address through graph-formed reasoning. Subsequently, we summarize related works into
two distinct types of reasoning: think on graphs and verify on graphs.
\begin{figure*}
    \centering
     \includegraphics[width=\textwidth]{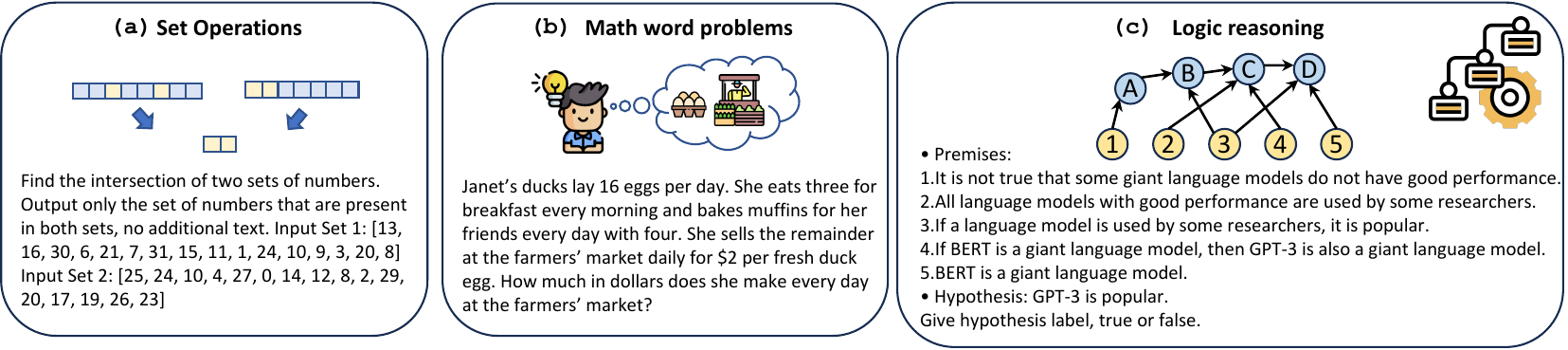} 
      \caption{Three examples of graph-formed reasoning tasks.}
      \label{graph-form-reasoning} 
\end{figure*}
\subsection{Task Introduction}
LLM can solve more difficult problems through graph-formed reasoning, such as algorithmic problems, logical reasoning problems, and mathematical word problems, as shown in Figure~\ref{graph-form-reasoning}. Below, we detail three tasks that can be effectively addressed through the LLM graph-formed reasoning.
\\
\textbf{Set operations} mainly focus on set intersection. Specifically, the second input set is split into subsets and the intersection of those subsets with the first input set is determined with the help of the LLM, as shown in Figure~\ref{graph-form-reasoning} (a).
\\
\textbf{Math word problems} include single- and multi-step problems with addition, multiplication, subtraction, division, and other math operations. LLMs are required to understand mathematical relationships and then perform a multi-step reasoning process to ultimately reach an answer, as shown in Figure \ref{graph-form-reasoning} (b).
\\
\textbf{Logic reasoning} is a process aimed at inference and argumentation rigorously. LLMs start from a set of premises and reason towards a conclusion supported by those premises. For example, propositional logic consists of p, q, r, and various operations, as shown in Figure~\ref{graph-form-reasoning} (c).

Other tasks can be effectively solved through graph-formed reasoning, including document merging, multi-hop question answering, medical question answering and causal inference.

\begin{figure*}
    \centering
     \includegraphics[width=0.9\textwidth]{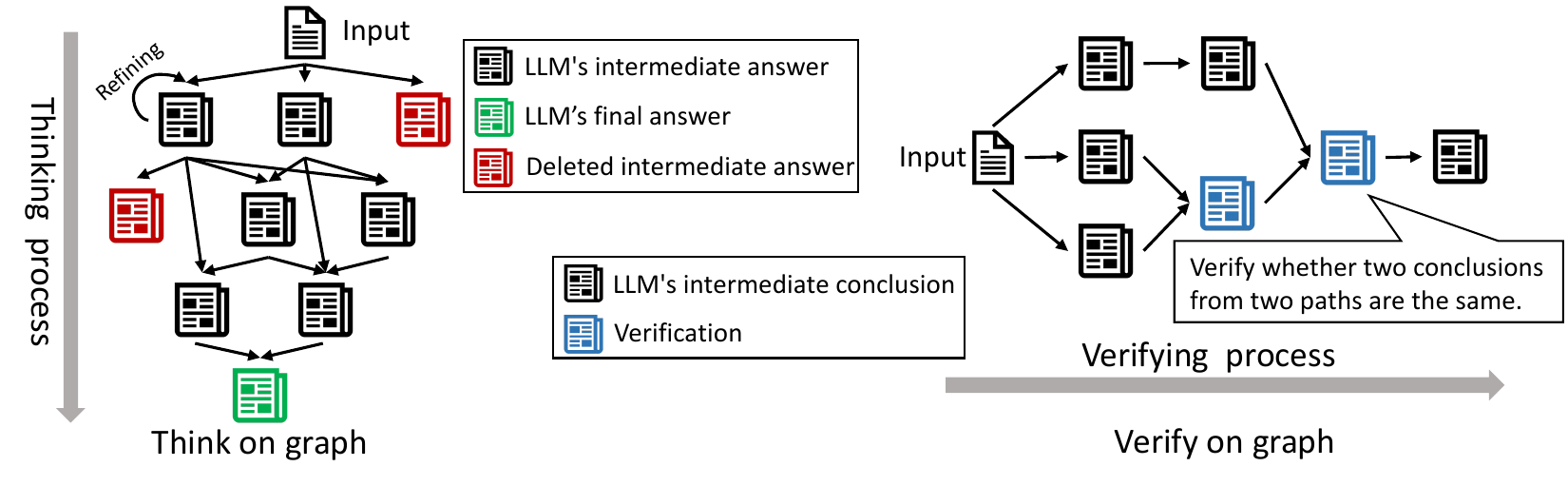} 
      \caption{Graph-formed reasoning: ``Think on graphs" involves reasoning over graph structures to derive conclusions, while ``Verify on graphs" entails using graph structures to validate the correctness of LLM outputs.}
      \label{graph-reasoning} 
\end{figure*}

\subsection{Graph-formed Reasoning Methods}
The graph form, with its inherent structural features, not only mimics human reasoning patterns but also validates answers from LLM through the relationships between nodes and local structure. Existing work can roughly be divided into two categories: \emph{\textbf{think on the graph}} and \emph{\textbf{verify on the graph}}, as shown in Figure \ref{graph-reasoning}. Think on the graph refers to LLM thinking in the form of a graph, where each node on the graph represents a step in the thinking process or an intermediate conclusion during thinking, and the edges on the graph indicate the direction of LLM inference or the relationships between intermediate thinking steps. In this way, the LLM thinking process can be visually represented in graph form. Verify on the graph means verifying the consistency and correctness of answers by utilizing the graph's structure. For example, if the end node of different paths is the same, the results derived from different paths should be the same. If contradictory conclusions arise, then the obtained conclusion is incorrect.

\subsubsection{\textbf{Think on the graph}}
The GoT* reasoning method~\cite{yao2023beyond} is proposed with a two-stage framework to enable LLM to reason on a graph for answering multiple-choice questions by converting the input query into a graph.
Although GoT* allows LLM to enhance the graph using multimodal information, it does not reason for step-by-step deduction in graph form. The Graph of Thought (GoT) \cite{besta2024graph} represents LLM's intermediate thinking as an arbitrary graph, facilitating powerful prompting for solving algorithmic problems like sorting and keyword counts. LLM thoughts are depicted as vertices with edges representing dependencies between them. 
Existing works also explore significant components within LLM-generated causal graphs. {Vashishtha et al. \cite{vashishtha2023causal} utilize} the topological ordering of variables in a graph to enhance the causal inference capabilities of LLMs, thereby determining the causal relationships among different variables. 
For practical applications, MindMap~\cite{wen2024mindmap} extracts and merges subgraphs from existing KGs. LLM then infers disease and treatments over these subgraphs to derive the final answer.
Extending beyond the capabilities of a single LLM, multiple LLMs can also be collaboratively harnessed to tackle complex mathematical challenges. CR~\cite{zhang2023cumulative} is proposed as a more human-like reasoning process. CR utilizes three LLMs in different roles: the proposer, verifier, and reporter. 

\subsubsection{\textbf{Verify on the graph}}
\textit{Verify on the graph} is to validate the intermediate reasoning results of LLM to enhance its performance. GraphReason~\cite{cao2023enhancing} assumes a logical connection between the intermediate steps of different inference paths created by LLM. 
However, this work trains an extra model to verify the correctness of the graph formed by LLM-generated solutions rather than utilizing the relationships between nodes in the graph. The Graph-guided CoT~\cite{park2023graph} evaluates whether the generated rationales by LLMs can answer the original question based on the consistency of adjacent rationales in the graph, 
Numerous cutting-edge studies have focused on quantifying the trustworthiness of LLMs and assessing the reliability of their responses. D-UE~\cite{da2024llm} computes a final uncertainty measurement on a constructed directed graph, which evaluates the trustworthiness of the LLM's responses.

\subsection{Summary of Methods, Challenges and Future Directions}
{In summary, graph-formed reasoning is categorized into \textbf{\textit{think on the graph}} and \textbf{\textit{verify on the graph}}. \textbf{Think on the graph} refers to using the graph structure to derive the final conclusion during the reasoning process with LLM. \textbf{Verify on the graph} involves treating the intermediate or final results generated by LLM as nodes on the graph and using the graph to determine if there are contradictions between the nodes, thus verifying the correctness of the LLM output.}

{For ``think on the graph", a common issue with existing approaches is their lack of convenience. Compared to CoT and self-consistency prompting techniques, the reasoning processes in current works are complex, requiring multiple stages of reasoning and validation. Graph of thought methods are not plug and play, which contradicts the original intent of prompts. Even though using more LLMs can simplify the reasoning and validation process, it raises the cost and barrier to entry for reasoning. Therefore, the current challenge is to find a plug-and-play, low-barrier LLM graph reasoning method that improves LLM reasoning capabilities.
For ``verify on the graph", the current approaches have yet to utilize the nature of the graph structure for validation. Existing methods either retrain a model to determine correctness or use a KG for assessment without using the relationships between nodes to infer whether the conclusions within each node in the graph are correct.}

{The future direction of ``think on the graph" could focus on developing a plug-and-play, low-barrier LLM graph reasoning method that enhances LLM reasoning abilities, a pressing issue that needs to be addressed. On the other hand, the future direction of ``verify on the graph" could explore how to utilize the relationships between nodes in the graph structure to verify the outputs of LLM or the reasoning process itself.}

\section{Graph representation}
\label{representation section}
\begin{figure*}
    \centering
    \includegraphics[width=\textwidth]{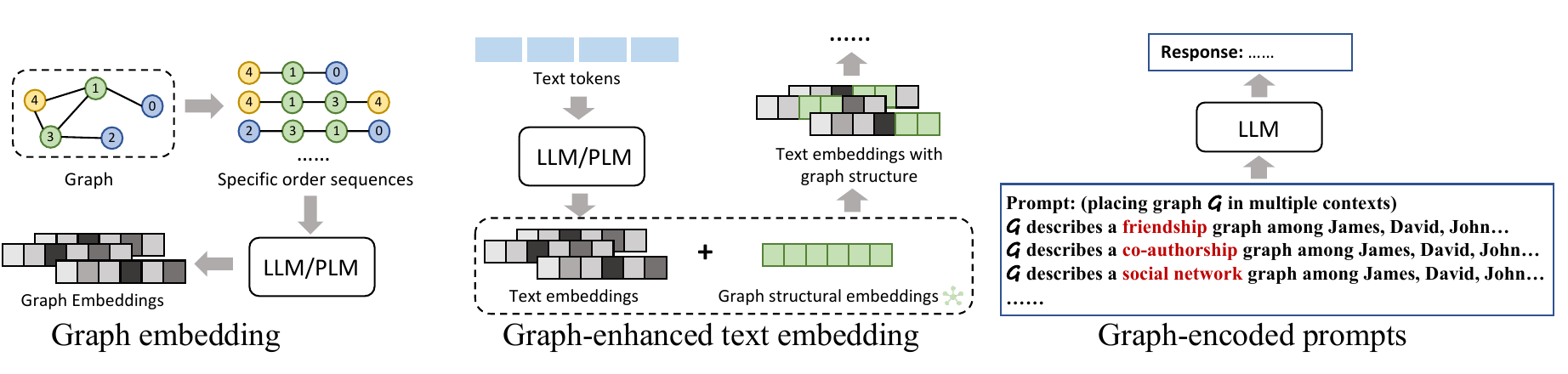} 
      \caption{Three types of graph representation: graph embedding, graph-enhanced text embedding, and graph-encoded prompts. }
      \label{graph-representation} 
\end{figure*} 

\subsection{Task introduction}
LLMs' powerful text representation abilities empower text embeddings to capture deeper semantic nuances, which also can enhance graph representations, particularly for Text Attributed Graphs (TAGs). {When dealing with structured text data, the key challenge is integrating graph structures into text embeddings produced by LLMs to enhance their informativeness or enable LLMs to process text embeddings with graph structures within the text space. Moreover, effectively incorporating the graph description within the prompt is essential for LLMs, especially in closed-source models like ChatGPT, where the embedding is invisible. How the graph is encoded within the prompt influences the model's comprehension of the graph.} Thus, we summarize the following three types of graph representation: \emph{graph embedding}, \emph{graph-enhanced text embedding}, and \emph{graph-encoded prompts}, as shown in Figure \ref{graph-representation}. 

\subsubsection{\textbf{Graph embedding}}
Graph embedding focuses on transforming a graph into a specific ordered sequence, which is then fed into an LLM/PLM to learn the token embedding within sequences using their semantic capturing ability, and then derive the graph embedding based on the token embedding. 

\subsubsection{\textbf{Graph-enhanced text embedding}}
Graph-enhanced text embedding emphasizes incorporating structural embedding into text embedding. There are two types of embeddings: structural embedding, which captures the local structure, and text embedding, which captures the semantic meaning. How to combine these two types of embeddings is the core of graph-enhanced text embedding.

\subsubsection{\textbf{Graph-encoded prompts}}
Graph-encoded prompts concentrate on how to describe a graph so that LLMs can understand it more efficiently and then input it into LLMs. For instance, in a regular graph, the graph can be placed in a story context by assuming that the relationships between the nodes are friends or colleagues.


\subsection{Graph Representation Methods}
For each of three graph representations mentioned above, we present their specific objectives and techniques as follows.

\subsubsection{\textbf{Graph embedding}}
Texts are sequential data, while graph data is structural, posing a challenge for LLMs. LLMs excel at handling texts but struggle with graphs. How do LLMs effectively process graphs and derive effective graph representations? Graph embedding methods use specific order sequences to represent the graph, where specific order represents graph structure. WalkLM~\cite{tan2024walklm} aims to enhance graph representations in TAGs by utilizing a small language model. Initially, text sequences are generated on TAGs through random walks, capturing graph structural features. Subsequently, these sequences are input into a masked language model for the self-supervised training process, leading to the integration of semantic and structural graph embeddings. To adapt LLM to effectively derive semantic-enhanced graph embeddings, Path-LLM\cite{shang2024path} proposes a new path selection mechanism, aligning with casual language modeling LLMs. Path-LLM first selects long shortest paths covering bridge paths between different dense groups and then convert long shortest paths into L2SP-based shortest paths in designed ways. Path-LLM uses L2SP-based shortest paths for LLM self-supervised generation process to derive enhanced graph embeddings. 
Existing methods also transform graphs into the natural language based on grammar trees. GraphText~\cite{zhao2024graphtext} reformulates graph reasoning as text-to-text problems, establishing text input and output spaces. GraphText first constructs grammar trees for graphs, then traverses them to generate graph text sequences, and finally maps the graph to the text space. The text input is then fed into an LLM, with the LLM results mapped to the label space, effectively enabling LLMs to handle graph tasks.
While the methods mentioned above primarily concentrate on textual data, there is a growing emphasis on utilizing multimodal data to capture more comprehensive information. GALLON~\cite{xu2024llm} integrates the strengths of LLMs and GNNs by distilling multimodal knowledge into a unified MLP. This approach combines the rich textual and visual data of molecules with the structural analysis capabilities of GNNs.

\begin{figure}
    \centering
    \includegraphics[width=0.48\textwidth]{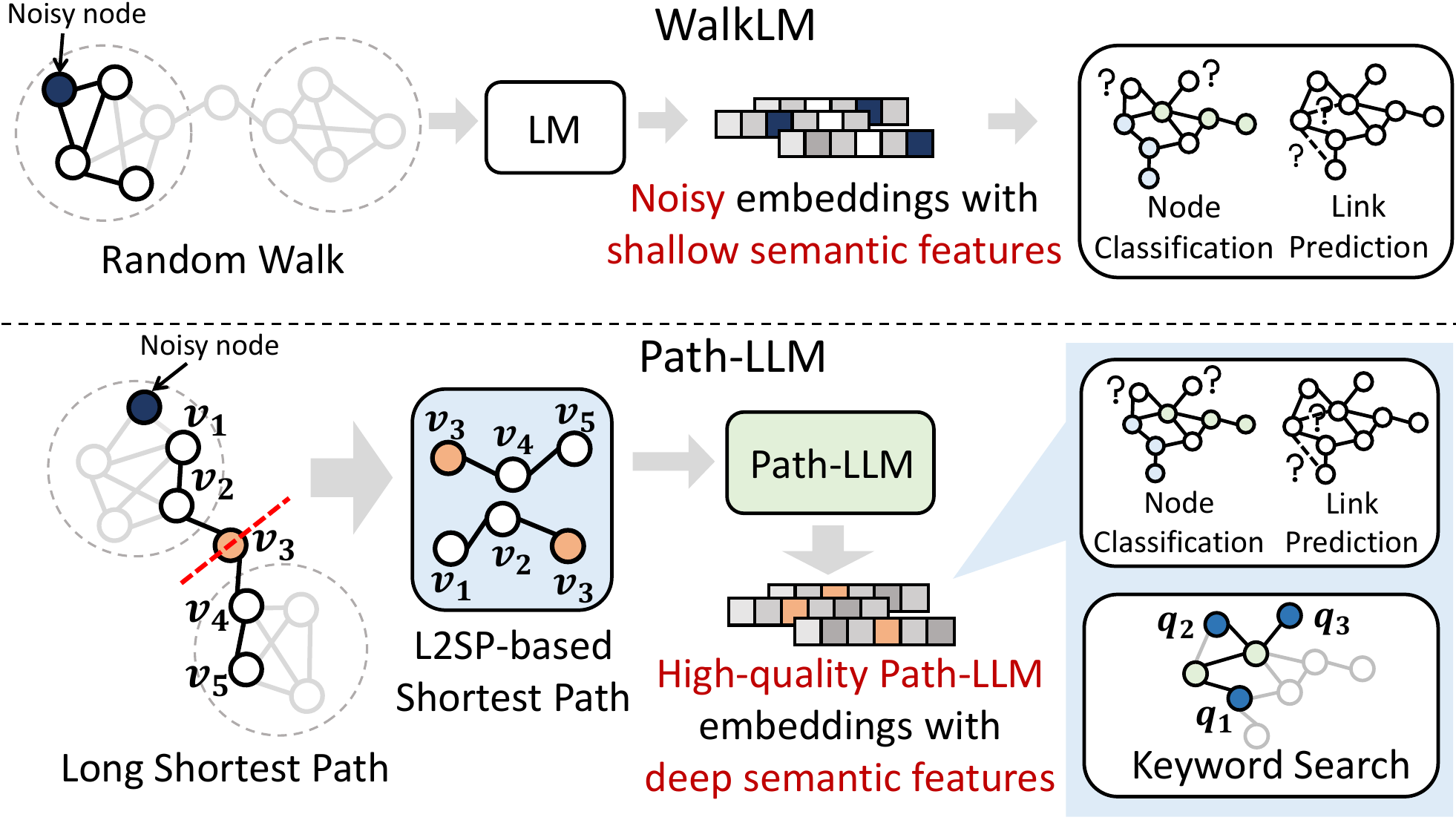} 
      \caption{
      WalkLM~\cite{tan2024walklm} v.s. Path-LLM~\cite{shang2024path}.
      }
      \label{walklm-vs-pathllm} 
\end{figure}

\stitle{Discussions of Path-LLM~\cite{shang2024path} vs. WalkLM~\cite{tan2024walklm}.}
This study compares two methods that utilize language models through unsupervised training to generate unified graph embeddings: WalkLM and Path-LLM. WalkLM emphasizes small language models employing the masked language modeling strategy, while Path-LLM concentrates on LLMs with the causal language modeling process, which is the dominant strategy for LLMs. Both methods extract graph structural features as self-supervised signals. As shown in Figure~\ref{walklm-vs-pathllm}, WalkLM uses random walks to represent graph structural features for language model learning. In contrast, Path-LLM introduces a novel L2SP-based shortest paths selection mechanism that mitigates noise, capturing relationships between different dense groups as well as features within dense groups. Furthermore, Path-LLM aligns the order of nodes in designed L2SP-based paths with the direction of the causal language modeling process,
ensuring that LLMs can effectively learn the graph structure. 

\subsubsection{\textbf{Graph-enhanced text embedding}} {Graph-enhanced text embedding emphasizes adding graph structural embeddings as additional information into text embeddings.} DGTL~\cite{qin2023disentangled} adds graph structure embeddings into text embeddings for node classification, where LLMs derive text embeddings. 
While DGTL~\cite{qin2023disentangled} concentrates on utilizing LLMs to integrate text and graph structure for graph tasks, G2P2~\cite{wen2023augmenting} emphasizes merging graph structure with text to address text classification tasks. Textual data commonly exhibits network structures, such as hyperlinks in purchase networks, which hold semantic relationships that can enhance text classification performance. 
\subsubsection{\textbf{Graph-encoded prompts}}
{Current related works focus on describing graph structures through natural language to LLMs in prompts without deeply learning the graph structure, which can lead to an LLM's insufficient understanding of complex structural relationships.}
{Therefore, effectively encoding graphs in the prompt is vital for LLMs to comprehend graph structure and solve graph tasks. Graph encoding refers to how graphs are represented in the textual prompt.}
TalkGraph~\cite{fatemi2024talk} introduces diverse graph encoding techniques by placing the same graph in multiple contexts. This strategy highlights how a node can be interpreted differently based on the context. For instance, a node could represent a person named David, with edges indicating various relationships like co-authorships or friendships. When asking LLM the degree of one node, in the given contexts, that equals how many friendships David has.
In contrast, TalkGraph~\cite{fatemi2024talk} primarily emphasizes text modality graph encoding, while GraphTMI~\cite{das2024modality} employs three encoding modalities, namely text, image, and motif, to encode graphs. 
Specifically, the text modality encoding provides insights into local structures, while the motif modality encoding captures essential graph patterns like stars, triangles, and cliques, offering a balanced perspective on local and global information. 
In comparing these two methods, TalkGraph~\cite{fatemi2024talk} focuses on diverse graph encoding within text modality by constructing contexts, whereas GraphTMI~\cite{das2024modality}  utilizes multiple modalities to encode graphs comprehensively, enhancing the LLMs' ability to understand graph structures.

\subsection{Summary of Methods, Challenges and Future Directions}
{In summary, graph embedding focuses on transforming a graph into a specific ordered sequence, which is then fed into an LLM to learn the sequence's embedding and derive the graph embedding. On the other hand, graph-enhanced text embedding emphasizes incorporating structural embedding into text embedding. Lastly, graph-encoded prompts concentrate on how to describe a graph and input it into an LLM.
However, due to LLMs' powerful text representation capabilities, the first two methods exhibit a deep semantic understanding of graph attributes. However, they still need suitable structural information capturing, which remains rudimentary and inadequate. Additionally, aligning the graph structure features with text features to better represent the graph's features is a current issue that needs to be addressed.
For graph-encoded prompts, most methods build a narrative context for the graph or describe it multimodally before feeding it into an LLM. Both methods enable the LLM to interpret the graph from various perspectives to improve performance. The critical challenge currently lies in designing diverse and easily understandable graph descriptions for LLMs, conveying essential graph descriptions while enhancing the LLM's comprehension of the input description.}

\stitle{Discussions of unsupervised vs. supervised graph representation methods} {Current methodologies for LLM-based graph representation can be classified into two main categories: supervised and unsupervised graph representation methods. In this paper, due to the self-supervised learning process of LLMs not leveraging supervised signals from downstream tasks, we categorize methods that train LLMs through a self-supervised way to derive unified graph representation as unsupervised methods.
Numerous cutting-edge supervised techniques are specifically designed for single graph learning tasks, such as node classification, exemplified by TAPE~\cite{he2023harnessing} and GraphAdapter~\cite{huang2024can2}. Furthermore, further supervised approaches train their models across various tasks, such as OFA~\cite{liu2024one} and LLaGA~\cite{chen2024llaga}.
Unsupervised methods leverage graph structural features as self-supervised signals to better understand the inherent nature of graphs. These methods derive unified graph embeddings that can be applied across various downstream tasks. Unlike supervised methods, which require extensive labeled data and are specifically trained for particular downstream tasks, unsupervised methods do not need labeled data or human manual resources. Consequently, unsupervised methods can be applied to a broader range of downstream tasks, including graph data analytics and even NP-hard problems such as keyword search~\cite{shang2024path}. While supervised methods typically demonstrate superior performance on specific downstream tasks, the flexibility and reduced resource requirements of unsupervised methods make them a valuable and empowering alternative in many situations.}
\section{Graph-LLM-based applications}
\label{application section}

Graph-LLM-based applications refer to frameworks that integrate graphs with LLMs. Apart from their applications in graph-related tasks, they are also utilized in various other domains, such as recommendation systems.
Common frameworks involve combining GNNs with LLMs, merging graph data with LLMs, and exploring additional innovative approaches that leverage the advantages of graph structures and language models for diverse applications.

\subsubsection{\textbf{Task planning}} {Graph structures are utilized to encode the world knowledge for LLM-based agents for task planning in the open world~\cite{wu2024can}.
Optimus-1~\cite{li2024optimus} transforms knowledge into a hierarchical directed graph for LLMs that allows agents to explicitly represent and learn world knowledge for long-horizon tasks in an open world. SayPlan~\cite{rana2023sayplan} conducts a semantic search of task-relevant 3D scene graphs to encode world model information to the LLM for scaling task planning for robots to large, multi-room environments.}

\subsubsection{\textbf{Recommendation systems}}
LLMs have been extensively utilized in recommendation systems~\cite{wu2024exploring}.
Graph structures are crucial for recommendation systems, as many tasks involve learning from user-item interaction networks. Sensitive domains such as educational recommendations require minimizing the risk of hallucinations in LLMs, 
combining with KGs~\cite{abu2024knowledge}. 
LLMHG~\cite{chu2024llm} facilitates a human-centric modeling of user preferences by integrating the reasoning capabilities of LLMs with the structural advantages of hypergraph neural networks.

\subsubsection{\textbf{Conversational understanding}} 
By combining LLM with graph traversal, collaborative query rewriting~\cite{chen2023graph} is proposed to improve the coverage of unseen interactions, addressing the flawed queries users pose in dialogue systems. Flawed queries often arise due to ambiguities or inaccuracies in automatic speech recognition and natural language understanding. When integrated with graph traversal, LLM can effectively navigate through the graph structure to retrieve relevant information and provide more accurate responses. 

\subsubsection{\textbf{Response forecasting}}  
LLM can effectively handle social networks and extract latent personas from users' profiles and historical posts. SOCIALSENSE~\cite{sun2023decoding} is proposed to utilize LLMs to extract information to predict the reactions of news media. By analyzing individuals' characteristics and behavior patterns within social networks, LLM can effectively predict the impact of news releases and prevent unintended adverse outcomes.



{Many other applications leverage LLMs and graphs across various domains. MANAGER~\cite{ouyang2024modal} utilizes LLMs with multimodal information and the dynamic financial knowledge graph (FinDKG) to address financial prediction. GRACE~\cite{lu2024grace} incorporates graph structural information in the code and in-context learning to empower LLM-based software vulnerability detection. 
Additionally, Li et al. \cite{li2024llm2} employ LLMs to transform input scene graphs into affordance-enhanced graphs, thereby facilitating household rearrangement. 
Furthermore, GPT4GNAS~\cite{wang2023graph} leverages GPT-4 to generate graph neural network structures.}  
\section{Benchmark Datasets and Evaluations}
\label{benchmark and evaluation}

\begin{table*}[h!t]
\scriptsize
   \centering
   \renewcommand\arraystretch{1.2}
   \caption{A summary of LLM-GGA representative methods with datasets and source links.}
   \begin{tabular}{lp{4.5cm}p{4cm}ll}\hline
    Method&      Dataset&  LLM &     Task & Link \\ \hline
    InstrucGLM\cite{ye2024language} &   ogbn-arxiv, Cora, PubMed & Flan-T5, Llama-v1-7b  &   Node, link & \href{https://github.com/agiresearch/InstructGLM}{code link}\\
    GraphAdapter\cite{huang2024can2} &   ogbn-arxiv, Instagram, Reddit & Llama2-13B & Node & \href{https://github.com/zjunet/GraphAdapter}{code link}  \\
    GPT4Graph\cite{guo2023gpt4graph} &  ogbn-arxiv,Aminer,Wiki,MetaQA & InstructGPT-3  & Graph structure, node, graph & \href{https://anonymous.4open.science/r/GPT4Graph}{code link}\\
    LLMtoGraph\cite{liu2023evaluating} & Synthetic data & GPT-3.5, GPT-4, Vicuna, Lazarus & Multi-hop Reasoning & \href{https://github.com/Ayame1006/LLMtoGraph}{code link} \\ 
    Graph-LLM\cite{chen2024exploring} & ogbn-arxiv, Cora, PubMed, ogbn-products & LLaMA, Palm &  Node & \href{https://github.com/CurryTang/Graph-LLM}{code link}\\
    TAPE\cite{he2023harnessing} &  ogbn-arxiv, Cora, PubMed, ogbn-products & GPT-3.5 &  Node & \href{https://github.com/XiaoxinHe/TAPE}{code link}\\
    LLM4DyG\cite{zhang2024llm4dyg}    &  LLM4DyG & GPT-3.5, Vicuna, Llama2, CodeLlama  &  Graph & \href{https://github.com/wondergo2017/LLM4DyG}{code link}\\
    GraphGPT\cite{tang2024graphgpt}   &   ogbn-arxiv, Cora, PubMed & Vicuna & Node & \href{https://github.com/HKUDS/GraphGPT}{code link}\\  
    Graph-ToolFormer\cite{zhang2023graph}   &  GPR, Cora, Pubmed, Citeseer, Proteins, etc.
    & GPT-J  & Graph Q\&A, graph structure & \href{https://github.com/jwzhanggy/Graph_Toolformer}{code link}\\
    LLaGA \cite{chen2024llaga}  &  ogbn-arxiv, ogbn-products, Pubmed, Cora & GPT-3.5-turbo  & Node & \href{https://github.com/VITA-Group/LLaGA}{code link}\\  
    LLM-GNN \cite{chen2024label}  & Cora, Citeseer, PubMed, Wiki, ogbn-arxiv,etc.
    & GPT-3.5-turbo  & Node  & \href{https://github.com/CurryTang/LLMGNN}{code link}\\
    GraphTMI \cite{das2024modality} & Cora, Citeseer, Pubmed,GraphTMI & GPT-4, GPT-4V  &  Representation, node & \href{https://github.com/minnesotanlp/GraphLLM}{code link}\\
    WalkLM \cite{tan2024walklm}  &  PubMed, MIMIC-III, Freebase, FB15K-237 & PLMs  & Representation, node, link & \href{https://github.com/Melinda315/WalkLM}{code link}\\
    GraphText \cite{zhao2024graphtext}  &  Cora, Citeseer, Texas, Wisconsin, Cornell  & Llama2   & Node & \href{https://github.com/AndyJZhao/GraphText}{code link} \\
    TALK LIKE A GRAPH \cite{fatemi2024talk} & GraphQA &  PaLM  & Node, link & \href{https://github.com/google-research/talk-like-a-graph}{code link}\\
    Graph-guided CoT \cite{park2023graph} & Wiki, MusiQue, Bamboogle & Llama2  & Multi-hop question answering & -\\
    NLGraph \cite{wang2024can}  &  NLGraph & GPT-3.5, GPT-4  &  Link,node,graph,path,pattern & \href{https://github.com/Arthur-Heng/NLGraph}{code link} \\
    Gorilla~\cite{patil2024gorilla} &  APIBench  & Llama2, GPT-3.5, GPT-4  &  Multimodal graphs  & \href{https://gorilla.cs.berkeley.edu/}{code link}\\
    Collaborative Query Rewriting \cite{chen2023graph} & opportunity test sets, guardrail test set & Dolly V2  & Conversational understanding & - \\
    WHEN AND WHY \cite{huang2024can}  &  ogbn-arxiv, Cora, PubMed, ogbn-products &    ChatGPT  &   Node & \href{https://github.com/TRAIS-Lab/LLM-Structured-Data}{code link} \\
    CR \cite{zhang2023cumulative}  &  Folio, LogiQA, Proofwriter, Logicaldeduction & GPT-3.5-turbo, GPT-4, Llama2  &  Logic reasoning  & \href{https://github.com/iiis-ai/cumulative-reasoning}{code link} \\
    SOCIALSENSE \cite{sun2023decoding}  &  RFPN, Twitter & PLMs  &   Response forecasting & \href{https://github.com/chenkaisun/SocialSense}{code link}  \\ 
    MindMap \cite{wen2024mindmap} & GenMedGPT-5k, CMCQA, ExplainCPE & GPT-3.5, GPT-4 & Medical Q\&A & \href{https://github.com/wyl-willing/MindMap}{code link} \\
    PiVe~\cite{han2024pive}  & KELM, WebNLG+2020, GenWiki  & GPT-4   & Graph generation  &  \href{https://github.com/Jiuzhouh/PiVe}{code link}  \\
    Graph of Thought(GoT) \cite{besta2024graph}   &  Non-open source data &  GPT3.5  &  Graph-formed reasoning  & \href{https://github.com/spcl/graph-of-thoughts}{code link}  \\
    GLEM\cite{zhao2023learning}  & ogbn-arxiv, ogbn-products, ogbn-papers & PLMs  &  Node  & \href{https://github.com/AndyJZhao/GLEM}{code link}  \\
    LPNL\cite{bi2024lpnl} & OAG  & T5-base & Link &  - \\
    SIMTEG\cite{duan2023simteg} & ogbn-arxiv, ogbn-products, ogbl-citation2 & PLMs  & Node, link & \href{https://github.com/vermouthdky/SimTeG}{code link} \\
    Llmrec \cite{wei2024llmrec} & Netflix, MovieLens   & gpt-3.5-turbo-16k   & Recommendation  & \href{https://github.com/HKUDS/LLMRec}{code link} \\
    OFA \cite{liu2024one} & ogbn-arxiv, Cora  & PLMs   &  Node, link, graph  &  \href{https://github.com/LechengKong/OneForAll}{code link}  \\  
    GraphTranslator \cite{zhang2024graphtranslator}  &  ogbn-arxiv, Taobao  &  GPT-3.5, GPT-4   &  Node, graph Q\&A &  \href{https://github.com/alibaba/GraphTranslator}{code link} \\
    GAugLLM~\cite{fang2024gaugllm} &  PubMed, ogbn-arxiv, Amazon & Llama2  &  Node  & \href{https://github.com/NYUSHCS/GAugLLM}{code link}   \\
    GPT4GNAS\cite{wang2023graph}  &   ogbn-arxiv, Cora, PubMed, Citeseer  &  GPT-4   &  Graph neural architecture search   &   -   \\
    Graphllm\cite{chai2023graphllm}  &  NLGraph  & Llama2-7B, Llama2-13B   &  Link, node, graph, path, pattern &  \href{https://github.com/mistyreed63849/Graph-LLM}{code link}   \\
    G2P2 \cite{wen2023augmenting} &  Cora, Amazon   &   PLMs   &  Representation   &   \href{https://github.com/WenZhihao666/G2P2}{code link}   \\
    ChatGraph\cite{peng2024chatgraph} &  Gradio  &  GPT-4V, Next-GPT    &  Link, node, graph, application    &  - \\
    GoT*\cite{yao2023beyond} &  AQUA-RAT, ScienceQA   &  T5-base   &  Graph-formed reasoning   &  \href{https://github.com/Zoeyyao27/Graph-of-Thought}{code link}   \\
    KGP\cite{wang2024knowledge}  &  HotpotQA, IIRC, Wiki, MuSiQue, PDFTriage  &  Llama2  &  KG+LLM  &   \href{https://github.com/YuWVandy/KG-LLM-MDQA}{code link}   \\
    Head-to-Tail \cite{sun2024head} &   DBpredia, Movie, Book, Academics    &  GPT-4    &  KG+LLM   &  \href{https://github.com/facebookresearch/head-to-tail}{code link}   \\
    GLaM\cite{dernbach2024glam}  &  DBLP, UMLS    &   Llama2-7B   &  KG+LLM &   -  \\
    ToG\cite{sun2024think}  &  CWQ, WebQSP, GrailQA, QALD10-en, etc.    &  GPT-3.5, GPT-4, Llama2   &  KG+LLM   &  \href{https://github.com/IDEA-FinAI/ToG.}{code link}  \\
    Autoalign\cite{zhang2023autoalign} &  DBpedia, Wikidata  & ChatGPT, Claude    & KG+LLM   &  \href{https://github.com/ruizhang-ai/AutoAlign}{code link}  \\  
    GNP\cite{tian2024graph} &  ConceptNet, UMLS, OpenBookQA, etc.    &  FLAN-T5   & KG+LLM   &  \href{https://github.com/meettyj/GNP}{code link}  \\
    RoG\cite{tian2024graph}  &  WebQSP, CWQ, Freebase    &  Llama2-7B     &    KG+LLM    &    \href{https://github.com/RManLuo/reasoning-on-graphs}{code link} \\
  \hline
    \end{tabular}
    \label{summary-methods}
\end{table*}

\subsection{Datasets}

Table~\ref{summary-methods} summarizes representative works in this survey, the popular and new datasets, the performed  LLM-GGA tasks, and links to the corresponding open-source codes.

\label{benchmark}
\subsubsection{\textbf{Popular datasets}}
Popular datasets refer to graph benchmarks that are widely and frequently used. We have systematically categorized these popular benchmarks in terms of {five} LLM-GGA directions as follows. 
\begin{itemize}
    \item \textbf{Graph structure understanding}: ogbn-arxiv\cite{hu2020open}, ogbn-products\cite{hu2020open}, Cora\cite{mccallum2000automating}, CiteSeer\cite{giles1998citeseer}, Aminer(DBLP)\cite{tang2008arnetminer}, MetaQA\cite{zhang2018variational}, Wikidata5M\cite{wang2021kepler}, PROTEINS \cite{borgwardt2005protein}, MUTAG\cite{debnath1991structure}, NCI1\cite{wale2008comparison}, PTC\cite{toivonen2003statistical}, Foursqure \cite {kong2013inferring}.
    \item \textbf{Knowledge graphs and LLMs}: CWQ\cite{yih2016value}, WebQSP\cite{talmor2018web}, Wikidata\cite{wang2021kepler}, GrailQA \cite{gu2021beyond}, QALD10-en \cite{perevalov2022qald}.
    \item \textbf{Graph learning}: ogbn-arxiv\cite{hu2020open}, ogbn-products\cite{hu2020open}, ogb-papers110M\cite{hu2020open}, ogb-citation2\cite{hu2020open}, Cora\cite{mccallum2000automating}, CiteSeer\cite{giles1998citeseer}, Amazon-items\cite{wan2018item}, PubMed\cite{ni2019justifying}, Reddit\cite{hamilton2017inductive}, CoraFull \cite{bojchevski2018deep}, Amazon\cite{shchur2018pitfalls}, PROTEINS \cite{borgwardt2005protein}, COX2 \cite{rossi2015network}, BZR \cite{rossi2015network}, OAG \cite{huang2020analysis}.
    \item \textbf{Graph-formed reasoning}: GSM8K \cite{cobbe2021training}, SVAMP\cite{patel2021nlp}, FOLIO\cite{han2024folio}, Bayesian networks\cite{scutari2015bayesian}, CMCQA\cite{xia2022medconqa}, GenMedGPT-5k \cite{li2023chatdoctor}.
    \item \textbf{Graph representation}: Cora\cite{mccallum2000automating}, CiteSeer\cite{giles1998citeseer}, Goodreads-books\cite{zhang2023effect}, PubMed\cite{ni2019justifying}, Amazon\cite{shchur2018pitfalls},  MIMIIC-III \cite {johnson2016mimic}, Freebase \cite{bollacker2008freebase}, FB15K-237 \cite{toutanova2015observed}.
\end{itemize}

\subsubsection{\textbf{New datasets}}

We share several newly released benchmarks to evaluate LLM-based structure understanding ability and their potential to solve graph problems in Table~\ref{new benchmark}.

\begin{table}[!t]
\scriptsize
   \centering
   \renewcommand\arraystretch{1.3}
   \caption{A summary of new datasets.}
   \label{new benchmark}
   \begin{tabular}{llp{5.5cm}}\hline
    New Benchmark&      Link & Applicable Tasks \\ \hline
    GPR \cite{zhang2023graph} &   \href{https://github.com/jwzhanggy/Graph_Toolformer/tree/main/data}{link} & Graph structure understanding\\
    GraphTMI \cite{das2024modality} &  \href{https://github.com/minnesotanlp/GraphLLM}{link} & Graph structure understanding, graph representation \\
    LLM4DyG \cite{zhang2024llm4dyg} & \href{https://github.com/wondergo2017/LLM4DyG}{link} & Graph structure understanding\\  
    GraphQA \cite{fatemi2024talk} & \href{https://github.com/google-research/talk-like-a-graph}{link} & Graph representation \\
    NLGraph \cite{wang2024can} &  \href{https://github.com/Arthur-Heng/NLGraph/tree/main/NLGraph}{link} & Graph structure understanding\\
    Head-to-Tail \cite{sun2024head}  &  \href{https://github.com/facebookresearch/head-to-tail}{link} & Graph learning \\
    CS-TAG \cite{yan2023comprehensive} & \href{https://github.com/sktsherlock/TAG-Benchmark}{link} & Graph structure understanding, learning, representation  \\
    NLGIFT \cite{zhang2024can} & \href{https://github.com/MatthewYZhang/NLGift}{link} & Graph structure understanding \\
  \hline
    \end{tabular}
\end{table}

\begin{itemize}
    \item GPR \cite{zhang2023graph} contains 37 particular connected graph instances generated by the Networkx toolkit, which include the “bull graph,” “wheel graph,” “lollipop graph,” etc. Graph instances have about 15 nodes and 28 links on average.
    \item GraphTMI \cite{das2024modality} is a multi-modality benchmark, containing the hierarchy of graphs, associated prompts, and encoding modalities. Based on 
    the count of motifs and
    homophily in graphs, graph tasks have easy, medium, and hard levels.
    \item LLM4DyG \cite{zhang2024llm4dyg} is to evaluate whether LLMs are capable of understanding spatial-temporal information on dynamic graphs. Nine dynamic graph tasks are designed to assess LLM abilities from spatial and temporal dimensions.
    \item GraphQA \cite{fatemi2024talk} comprises a set of diverse fundamental graph problems with more varied and realistic graph structures compared to previous studies in LLM research.
    \item NLGraph \cite{wang2024can} is to examine whether language models can reason with graphs and structures. 
    It contains eight graph structure understanding tasks with varying algorithmic difficulties, enabling fine-grained analysis. 
    \item Head-to-Tail \cite{sun2024head} aims to measure how knowledgeable LLMs are, consisting of 18K question-answer (QA) pairs regarding head, torso, and tail facts in terms of popularity.
    \item CS-TAG \cite{yan2023comprehensive} is a comprehensive and wide-ranging compilation of benchmark for TAGs, 
    ranging from citation networks to purchase graphs. The collection consists of eight distinct TAGs sourced from diverse domains.
    \item NLGIFT \cite{zhang2024can} aims to evaluate whether LLM 
    can generalize beyond semantic, numeric, and structural reasoning patterns in the synthetic training data. NLGIFT contains 37,000 problems in total and features five types of patterns.
\end{itemize}
\vspace{-0.4cm}


\subsection{Evaluation Metrics}
\label{evaluation}
Evaluation metrics are essential to determine how well LLMs perform their understanding of graphs and how effectively models combining graphs and LLMs perform on various tasks is vital. We summarize important and widely-used evaluation metrics for each LLM-GGA task, as shown in Table~\ref{metrics}.

\subsubsection{\textbf{Graph structure understanding task}}
Several metrics usually used include the accuracy, ROUGE\cite{lin2004rouge}, BLEU\cite{papineni2002bleu}, time cost, comprehension, correctness, fidelity, and rectification comprehension. 
Moreover, the comprehension, correctness, fidelity, and rectification comprehension are new metrics \cite{liu2023evaluating} used to evaluate LLM graph understanding ability and accuracy.

\subsubsection{\textbf{Graph learning task}}
Various metrics evaluate models' effectiveness, efficiency, and computational demands. For the effectiveness of models, metrics such as accuracy, macro-F1, mismatch rate, and denial rate~\cite{das2024modality} are considered. In terms of efficiency, metrics like training time and tuned parameters are assessed. To evaluate the computational cost,
the GPU occupancy and the token limit fraction are used. 

\subsubsection{\textbf{Graph-formed reasoning task}}
The effectiveness metrics of graph-formed reasoning tasks include the accuracy, number of errors and cost, F1-score, precision, and recall\cite{goutte2005probabilistic}. Meanwhile, the efficiency of the reasoning process is evaluated through metrics such as the Latency-Volume trade-off.

\subsubsection{\textbf{Graph representation}}
The effectiveness of graph representation is mainly evaluated by its downstream tasks. Metrics such as accuracy and F1-score are commonly used for node classification, while AUC and Hits@k are employed in link prediction. Additionally, visualization can be utilized for directly assessing the effectiveness of graph representations.

\subsubsection{\textbf{Knowledge graphs and LLMs}}
KG-related tasks typically involve question-answering tasks. Evaluation metrics commonly used include accuracy, precision, recall, F1-score, Hits@k\cite{kleinberg1999web}, EM\cite{iacus2012causal}, MSE, hallucination rate\cite{sun2024head} and for some generative tasks, human evaluation may also be utilized.

\begin{table}[!t]
\footnotesize
   \centering
   \setlength\tabcolsep{2pt}
   \renewcommand\arraystretch{1.3}
   \caption{Frequently Used Metrics.}
   \label{metrics}
   \begin{tabular}{lp{5.2cm}}\hline
    Tasks&     Frequently used metrics  \\ \hline
    Graph structure understanding &   Accuracy, rouge, bleu, time cost, comprehension, correctness, fidelity, rectification comprehension\\
    \hline
    Graph learning &  Accuracy, F1, mismatch rate, denial rate, training time, tuned parameters, GPU occupancy, token limit fraction   \\ \hline
    Graph-formed reasoning & Accuracy, number of errors and cost, F1-score, precision, recall, latency-volume trade-off\\   \hline
    Graph representation & Visualization, metrics of downstream tasks \\ \hline
    Knowledge graphs and LLMs &  Accuracy, precision, recall, F1-score, Hits@k, EM, MSE, hallucination rate \\
  \hline
    \end{tabular}
\end{table}
\section{Future Directions}
\label{future direction}
In the following, we discuss the remaining challenges and promising future directions of LLM-GGA.

\stitle{A. LLM-CGA on complex graph patterns and tasks}. Existing works address simple graph tasks and patterns, such as the shortest path, clustering coefficient, maximum flow, etc. However, how to leverage LLMs to address a broad range of NP-hard problems, such as community search and interactive graph problems, is still an open question. In addition, current graph learning tasks mainly focus on simple node, edge, and graph classification, neglecting complex graph learning problems, such as the diverse classification outcomes arising from homogeneous and heterogeneous graphs.

\stitle{B. LLM-CGA on complex and big graph data}. Due to the limited input length of LLM, the graph sizes inputted through prompts typically consist of dozens of nodes. However, for large graphs with tens of thousands of nodes and edges, how can LLMs with limited input length solve such large graphs? A larger input window is required in the case of attributed graphs, where both node and edge attributes need to be considered along with the graph structure. {Except for the large-scale property, big graph data usually has diverse types, including higher-order networks, multi-layer networks, uncertain networks, and dynamic graphs, where each type of graph contains unique and special properties. For instance, dynamic graphs can be represented as ordered lists or asynchronous streams of timed events, capturing patterns of temporal network evolution, such as the addition or removal of nodes and edges. In the future, LLM-based agents with the API calls are one potential direction to accurately handle large-scale and complex graphs. }

\stitle{C. Advanced Graph Prompts}. 
Graph prompts for LLMs have yet to be sufficiently explored. Relying solely on manual prompts and self-prompting has limited capabilities in improving model performance, as they only explore the existing abilities of LLM. As shown in Section \ref{structure prompt}, LLMs can be trained as agents to utilize tools for graph tasks that are hard to solve, like API call prompt\cite{zhang2023graph}. GoT \cite{zhao2023gimlet} is also a graph reasoning paradigm that enables LLMs to provide correct answers. {Future work based on the graph reasoning paradigm can consider cost-effective approaches for GoT, such as pruning and tricks to reduce algorithm complexity. In the future, it would be beneficial to explore simpler GoT paradigms that can improve the effectiveness of LLMs.}

\stitle{D. Graph foundation model and explainability}.
LLM is undoubtedly the foundational model in NLP. We can draw inspiration from LLMs to train a graph foundation model. {The current research~\cite{mao2024position} has primarily introduced graph foundation models in the form of graph-aware tuning LLMs and also GFM based on graph vocabulary.} Exploring graph foundation models is a future direction for graph tasks. In addition, we discuss the explainability.
LLMs' reasoning and thinking process can be step by step.
The combination of LLMs and GNNs has the potential to offer a more transparent approach to solving graph problems by leveraging the LLM's reasoning abilities. {If the combination of LLMs and GNNs is interpretable, it can be utilized for various tasks, including recommendation systems, drug discovery, and fraud detection. This combination can develop more reliable and efficient decision-making systems across multiple domains.}

\stitle{{E. Security and privacy}}. 
{Due to the dependencies on third-party vendors and service providers to carry out the LLM-related applications, data security and LLM architecture security increasingly become key issues for practical applications. For data security, knowledge graphs can be employed to efficiently store and retrieve information concerning privacy policies for LLMs, like PrivComp-KG~\cite{garza2024priv}, which is a potential direction in RAG domains. For LLM architecture security, several attack strategies can infer different parts of the graph data and LLM architectures, for example, model extraction attack, graph structure reconstruction, attribute inference attacks, and membership inference attacks~\cite{guan2024graph}. In the future, exploring the protection strategies for privacy data and LLM architectures can advance the reliable and practical LLM-based systems.}


\stitle{F. Graph databases and vector databases}. {Graph databases and vector databases are composed of embeddings to represent semi-structured data\cite{liu2025tigervector}\cite{khan2025graph}.  Owing to the strong representation abilities of LLMs, LLMs can generate more effective and context-aware node embeddings, thereby enhancing the retrieval accuracy of k-nearest neighbor algorithms (KNN) in graph databases and vector databases. One significant challenge lies in adapting LLMs to efficiently process large-scale databases. Future exploration can optimize the synergy between LLMs and vector databases, balancing both effectiveness and efficiency. }
\section{Conclusions}
\label{conclusions}
Large language models based generative graph analytics (LLM-GGA) has emerged as a promising direction to analyze classical graph tasks in an innovative way and tackle bottlenecks in analyzing complex graphs with text-attributes.  
This paper introduces a comprehensive structural taxonomy based on recent research, which classifies LLM-GGA research into three main directions: LLM-GQP, LLM-GIL, and graph-LLM-based applications. LLM-GQP encompasses graph understanding and knowledge graphs, while LLM-GIL involves graph learning, graph-formed reasoning, and graph representation. The motivation, challenges, and mainstream methods of each direction are thoroughly examined. This survey paper offers 40+ datasets, 
 a dozen of evaluation metrics, and the source codes for more than 40 mainstream methods in the discussed directions. It highlights the existing challenges in current methods and proposes several future directions to guide and motivate further research in this promising LLM-GGA field.

\bibliographystyle{IEEEtran}
\bibliography{ref}

\begin{thebibliography}{100}
\providecommand{\url}[1]{#1}
\csname url@samestyle\endcsname
\providecommand{\newblock}{\relax}
\providecommand{\bibinfo}[2]{#2}
\providecommand{\BIBentrySTDinterwordspacing}{\spaceskip=0pt\relax}
\providecommand{\BIBentryALTinterwordstretchfactor}{4}
\providecommand{\BIBentryALTinterwordspacing}{\spaceskip=\fontdimen2\font plus
\BIBentryALTinterwordstretchfactor\fontdimen3\font minus \fontdimen4\font\relax}
\providecommand{\BIBforeignlanguage}[2]{{%
\expandafter\ifx\csname l@#1\endcsname\relax
\typeout{** WARNING: IEEEtran.bst: No hyphenation pattern has been}%
\typeout{** loaded for the language `#1'. Using the pattern for}%
\typeout{** the default language instead.}%
\else
\language=\csname l@#1\endcsname
\fi
#2}}
\providecommand{\BIBdecl}{\relax}
\BIBdecl

\bibitem{wei2022finetuned}
J.~Wei, M.~Bosma, V.~Zhao, K.~Guu, A.~W. Yu, B.~Lester \emph{et~al.}, ``Finetuned language models are zero-shot learners,'' in \emph{ICLR}, 2022.

\bibitem{liu2024visual}
H.~Liu, C.~Li \emph{et~al.}, ``Visual instruction tuning,'' \emph{NeurIPS}, vol.~36, 2024.

\bibitem{rafailov2024direct}
R.~Rafailov, A.~Sharma, E.~Mitchell, C.~D. Manning, S.~Ermon, and C.~Finn, ``Direct preference optimization: Your language model is secretly a reward model,'' \emph{NeurIPS}, vol.~36, 2024.

\bibitem{huang2024trustllm}
Y.~Huang, L.~Sun, H.~Wang, S.~Wu, Q.~Zhang, Y.~Li \emph{et~al.}, ``Trustllm: Trustworthiness in large language models,'' in \emph{ICML}, 2024.

\bibitem{touvron2023llama}
H.~Touvron, L.~Martin, K.~Stone \emph{et~al.}, ``Llama 2: Open foundation and fine-tuned chat models,'' \emph{arXiv preprint arXiv:2307.09288}, 2023.

\bibitem{ouyang2022training}
L.~Ouyang, J.~Wu, X.~Jiang, D.~Almeida, C.~Wainwright \emph{et~al.}, ``Training language models to follow instructions with human feedback,'' \emph{NeurIPS}, vol.~35, pp. 27\,730--27\,744, 2022.

\bibitem{sun2025causalabstain}
Y.~Sun, A.~Zuo, W.~Gao, and J.~Ma, ``Causalabstain: Enhancing multilingual llms with causal reasoning for trustworthy abstention,'' \emph{arXiv preprint arXiv:2506.00519}, 2025.

\bibitem{zhuang2023toolqa}
Y.~Zhuang, Y.~Yu \emph{et~al.}, ``Toolqa: A dataset for llm question answering with external tools,'' \emph{NeurIPS}, vol.~36, pp. 50\,117--50\,143, 2023.

\bibitem{li2024flexkbqa}
Z.~Li, S.~Fan, Y.~Gu, X.~Li, Z.~Duan, B.~Dong \emph{et~al.}, ``Flexkbqa: A flexible llm-powered framework for few-shot knowledge base question answering,'' in \emph{AAAI}, vol.~38, no.~17, 2024, pp. 18\,608--18\,616.

\bibitem{zhang2023prompting}
B.~Zhang, B.~Haddow \emph{et~al.}, ``Prompting large language model for machine translation: A case study,'' in \emph{ICML}, 2023, pp. 41\,092--41\,110.

\bibitem{liu2024your}
J.~Liu, C.~S. Xia, Y.~Wang, and L.~Zhang, ``Is your code generated by chatgpt really correct? rigorous evaluation of large language models for code generation,'' \emph{NeurIPS}, vol.~36, 2024.

\bibitem{ni2023lever}
A.~Ni, S.~Iyer, D.~Radev, V.~Stoyanov, W.-t. Yih, S.~Wang, and X.~V. Lin, ``Lever: Learning to verify language-to-code generation with execution,'' in \emph{ICML}, 2023, pp. 26\,106--26\,128.

\bibitem{sen2008collective}
P.~Sen, G.~Namata, M.~Bilgic, L.~Getoor, B.~Galligher, and T.~Eliassi-Rad, ``Collective classification in network data,'' \emph{AI magazine}, vol.~29, no.~3, pp. 93--93, 2008.

\bibitem{hamilton2017inductive}
W.~Hamilton, Z.~Ying, and J.~Leskovec, ``Inductive representation learning on large graphs,'' \emph{NeurIPS}, vol.~30, 2017.

\bibitem{wu2018moleculenet}
Z.~Wu, B.~Ramsundar, E.~N. Feinberg, J.~Gomes, C.~Geniesse, A.~S. Pappu, K.~Leswing, and V.~Pande, ``Moleculenet: a benchmark for molecular machine learning,'' \emph{Chemical science}, vol.~9, no.~2, pp. 513--530, 2018.

\bibitem{broder2000graph}
A.~Broder, R.~Kumar, F.~Maghoul, P.~Raghavan, S.~Rajagopalan, R.~Stata, A.~Tomkins, and J.~Wiener, ``Graph structure in the web,'' \emph{Computer networks}, vol.~33, no. 1-6, pp. 309--320, 2000.

\bibitem{kipf2016semi}
T.~N. Kipf and M.~Welling, ``Semi-supervised classification with graph convolutional networks,'' in \emph{ICLR}, 2017.

\bibitem{velickovic2017graph}
P.~Velickovic, G.~Cucurull, A.~Casanova, A.~Romero \emph{et~al.}, ``Graph attention networks,'' \emph{stat}, vol. 1050, no.~20, pp. 10--48\,550, 2017.

\bibitem{gilmer2017neural}
J.~Gilmer, S.~S. Schoenholz, P.~F. Riley \emph{et~al.}, ``Neural message passing for quantum chemistry,'' in \emph{ICML}, 2017, pp. 1263--1272.

\bibitem{hong2009aggregation}
Y.~Hong, J.~W. Lam, and B.~Z. Tang, ``Aggregation-induced emission: phenomenon, mechanism and applications,'' \emph{Chemical communications}, no.~29, pp. 4332--4353, 2009.

\bibitem{cong2021provable}
W.~Cong, M.~Ramezani, and M.~Mahdavi, ``On provable benefits of depth in training graph convolutional networks,'' \emph{NeurIPS}, vol.~34, pp. 9936--9949, 2021.

\bibitem{fan2023generalizing}
S.~Fan, X.~Wang, C.~Shi, P.~Cui, and B.~Wang, ``Generalizing graph neural networks on out-of-distribution graphs,'' \emph{TPAMI}, 2023.

\bibitem{li2024survey}
Y.~Li, Z.~Li, P.~Wang, J.~Li, X.~Sun, H.~Cheng, and J.~X. Yu, ``A survey of graph meets large language model: progress and future directions,'' in \emph{IJCAI}, 2024, pp. 8123--8131.

\bibitem{ren2024survey}
X.~Ren, J.~Tang, D.~Yin, N.~Chawla, and C.~Huang, ``A survey of large language models for graphs,'' in \emph{SIGKDD}, 2024, pp. 6616--6626.

\bibitem{liu2024one}
H.~Liu, J.~Feng, L.~Kong, N.~Liang, D.~Tao, Y.~Chen, and M.~Zhang, ``One for all: Towards training one graph model for all classification tasks,'' in \emph{ICLR}, 2024.

\bibitem{jin2024large}
B.~Jin, G.~Liu, C.~Han, M.~Jiang, H.~Ji, and J.~Han, ``Large language models on graphs: A comprehensive survey,'' \emph{TKDE}, 2024.

\bibitem{liu2025graph}
J.~Liu, C.~Yang, Z.~Lu, J.~Chen, Y.~Li, M.~Zhang, T.~Bai, Y.~Fang, L.~Sun, P.~S. Yu \emph{et~al.}, ``Graph foundation models: Concepts, opportunities and challenges,'' \emph{TPAMI}, 2025.

\bibitem{li2024graph}
J.~Li, X.~Sun, Y.~Li, Z.~Li, H.~Cheng, and J.~X. Yu, ``Graph intelligence with large language models and prompt learning,'' in \emph{SIGKDD}, 2024, pp. 6545--6554.

\bibitem{peng2024graph}
B.~Peng, Y.~Zhu, Y.~Liu, X.~Bo, H.~Shi, C.~Hong, Y.~Zhang, and S.~Tang, ``Graph retrieval-augmented generation: A survey,'' \emph{arXiv preprint arXiv:2408.08921}, 2024.

\bibitem{han2024retrieval}
H.~Han, Y.~Wang, H.~Shomer, K.~Guo, J.~Ding, Y.~Lei, M.~Halappanavar, R.~A. Rossi \emph{et~al.}, ``Retrieval-augmented generation with graphs (graphrag),'' \emph{arXiv preprint arXiv:2501.00309}, 2024.

\bibitem{zhang2025survey}
Q.~Zhang, S.~Chen, Y.~Bei, Z.~Yuan, H.~Zhou, Z.~Hong, J.~Dong, H.~Chen, Y.~Chang, and X.~Huang, ``A survey of graph retrieval-augmented generation for customized large language models,'' \emph{arXiv preprint arXiv:2501.13958}, 2025.

\bibitem{lavrinovics2025knowledge}
E.~Lavrinovics, R.~Biswas, J.~Bjerva, and K.~Hose, ``Knowledge graphs, large language models, and hallucinations: An nlp perspective,'' \emph{Journal of Web Semantics}, vol.~85, p. 100844, 2025.

\bibitem{he2023harnessing}
X.~He, X.~Bresson, T.~Laurent, A.~Perold, Y.~LeCun, and B.~Hooi, ``Harnessing explanations: Llm-to-lm interpreter for enhanced text-attributed graph representation learning,'' in \emph{ICLR}, 2024.

\bibitem{yan2023comprehensive}
H.~Yan, C.~Li, R.~Long, C.~Yan, J.~Zhao, W.~Zhuang, J.~Yin, P.~Zhang \emph{et~al.}, ``A comprehensive study on text-attributed graphs: Benchmarking and rethinking,'' \emph{NeurIPS}, vol.~36, pp. 17\,238--17\,264, 2023.

\bibitem{zhao2023learning}
J.~Zhao, M.~Qu, C.~Li, H.~Yan, Q.~Liu \emph{et~al.}, ``Learning on large-scale text-attributed graphs via variational inference,'' in \emph{ICLR}, 2023.

\bibitem{fang2024gaugllm}
Y.~Fang, D.~Fan, D.~Zha, and Q.~Tan, ``Gaugllm: Improving graph contrastive learning for text-attributed graphs with large language models,'' in \emph{SIGKDD}, 2024, pp. 747--758.

\bibitem{tan2024walklm}
Y.~Tan, Z.~Zhou, H.~Lv, W.~Liu, and C.~Yang, ``Walklm: A uniform language model fine-tuning framework for attributed graph embedding,'' \emph{NeurIPS}, vol.~36, 2024.

\bibitem{choi2017gram}
E.~Choi, M.~T. Bahadori \emph{et~al.}, ``Gram: graph-based attention model for healthcare representation learning,'' in \emph{SIGKDD}, 2017, pp. 787--795.

\bibitem{li2022graph}
M.~M. Li, K.~Huang, and M.~Zitnik, ``Graph representation learning in biomedicine and healthcare,'' \emph{Nature Biomedical Engineering}, vol.~6, no.~12, pp. 1353--1369, 2022.

\bibitem{liao2018attributed}
L.~Liao, X.~He, H.~Zhang, and T.-S. Chua, ``Attributed social network embedding,'' \emph{TKDE}, vol.~30, no.~12, pp. 2257--2270, 2018.

\bibitem{campbell2013social}
W.~M. Campbell, C.~K. Dagli, and C.~J. Weinstein, ``Social network analysis with content and graphs,'' \emph{Lincoln Laboratory Journal}, vol.~20, no.~1, pp. 61--81, 2013.

\bibitem{li2022distilling}
Q.~Li, X.~Li, L.~Chen, and D.~Wu, ``Distilling knowledge on text graph for social media attribute inference,'' in \emph{SIGIR}, 2022, pp. 2024--2028.

\bibitem{guo2023gpt4graph}
J.~Guo, L.~Du, H.~Liu, M.~Zhou, X.~He, and S.~Han, ``Gpt4graph: Can large language models understand graph structured data? an empirical evaluation and benchmarking,'' \emph{arXiv preprint arXiv:2305.15066}, 2023.

\bibitem{liu2023evaluating}
C.~Liu and B.~Wu, ``Evaluating large language models on graphs: Performance insights and comparative analysis,'' \emph{arXiv preprint arXiv:2308.11224}, 2023.

\bibitem{zhang2024llm4dyg}
Z.~Zhang, X.~Wang, Z.~Zhang, H.~Li, Y.~Qin, and W.~Zhu, ``Llm4dyg: Can large language models solve spatial-temporal problems on dynamic graphs?'' in \emph{SIGKDD}, 2024, pp. 4350--4361.

\bibitem{huang2024can}
J.~Huang, X.~Zhang, Q.~Mei, and J.~Ma, ``Can llms effectively leverage graph structural information through prompts, and why?'' \emph{TMLR}, 2024.

\bibitem{wang2024can}
H.~Wang, S.~Feng, T.~He, Z.~Tan \emph{et~al.}, ``Can language models solve graph problems in natural language?'' \emph{NeurIPS}, vol.~36, 2024.

\bibitem{dong2023large}
Q.~Dong, L.~Dong, K.~Xu, G.~Zhou, Y.~Hao, Z.~Sui, and F.~Wei, ``Large language model for science: A study on p vs. np,'' \emph{arXiv preprint arXiv:2309.05689}, 2023.

\bibitem{fan2023nphardeval}
L.~Fan, W.~Hua, L.~Li, H.~Ling, Y.~Zhang, and L.~Hemphill, ``Nphardeval: Dynamic benchmark on reasoning ability of large language models via complexity classes,'' in \emph{ACL}, 2024.

\bibitem{dernbach2024glam}
S.~Dernbach \emph{et~al.}, ``Glam: Fine-tuning large language models for domain knowledge graph alignment via neighborhood partitioning and generative subgraph encoding,'' in \emph{AAAI}, vol.~3, no.~1, 2024, pp. 82--89.

\bibitem{sun2024think}
J.~Sun, C.~Xu, L.~Tang, S.~Wang, C.~Lin, Y.~Gong, L.~Ni, H.-Y. Shum, and J.~Guo, ``Think-on-graph: Deep and responsible reasoning of large language model on knowledge graph,'' in \emph{ICLR}, 2024.

\bibitem{tian2024graph}
Y.~Tian, H.~Song, Z.~Wang, H.~Wang, Z.~Hu, F.~Wang, N.~V. Chawla, and P.~Xu, ``Graph neural prompting with large language models,'' in \emph{AAAI}, vol.~38, no.~17, 2024, pp. 19\,080--19\,088.

\bibitem{luo2024reasoning}
L.~LUO, Y.-F. Li, R.~Haf, and S.~Pan, ``Reasoning on graphs: Faithful and interpretable large language model reasoning,'' in \emph{ICLR}, 2024.

\bibitem{xiong2024large}
S.~Xiong, A.~Payani, R.~Kompella, and F.~Fekri, ``Large language models can learn temporal reasoning,'' in \emph{Proceedings of the 62nd Annual Meeting of the Association for Computational Linguistics (Volume 1: Long Papers)}, 2024, pp. 10\,452--10\,470.

\bibitem{besta2024graph}
M.~Besta, N.~Blach, A.~Kubicek \emph{et~al.}, ``Graph of thoughts: Solving elaborate problems with large language models,'' in \emph{AAAI}, vol.~38, no.~16, 2024, pp. 17\,682--17\,690.

\bibitem{zhang2023cumulative}
Y.~Zhang, J.~Yang, Y.~Yuan, and A.~C.-C. Yao, ``Cumulative reasoning with large language models,'' \emph{arXiv preprint arXiv:2308.04371}, 2023.

\bibitem{yao2023beyond}
Y.~Yao, Z.~Li, and H.~Zhao, ``Beyond chain-of-thought, effective graph-of-thought reasoning in language models,'' \emph{arXiv preprint arXiv:2305.16582}, 2023.

\bibitem{zhao2024graphtext}
J.~Zhao, L.~Zhuo, Y.~Shen, M.~Qu, K.~Liu, M.~M. Bronstein, Z.~Zhu, and J.~Tang, ``Graphtext: Graph reasoning in text space,'' in \emph{AFM}, 2024.

\bibitem{qin2023disentangled}
Y.~Qin, X.~Wang, Z.~Zhang, and W.~Zhu, ``Disentangled representation learning with large language models for text-attributed graphs,'' \emph{arXiv preprint arXiv:2310.18152}, 2023.

\bibitem{shang2024path}
W.~Shang, X.~Zhu, and X.~Huang, ``Path-llm: A shortest-path-based llm learning for unified graph representation,'' \emph{arXiv preprint arXiv:2408.05456}, 2024.

\bibitem{wei2024llmrec}
W.~Wei, X.~Ren, J.~Tang, Q.~Wang, L.~Su, S.~Cheng, J.~Wang, D.~Yin, and C.~Huang, ``Llmrec: Large language models with graph augmentation for recommendation,'' in \emph{WSDM}, 2024, pp. 806--815.

\bibitem{wang2023graph}
H.~Wang, Y.~Gao, X.~Zheng, P.~Zhang, H.~Chen, J.~Bu, and P.~S. Yu, ``Graph neural architecture search with gpt-4,'' \emph{arXiv preprint arXiv:2310.01436}, 2023.

\bibitem{wu2024exploring}
L.~Wu, Z.~Qiu, Z.~Zheng, H.~Zhu, and E.~Chen, ``Exploring large language model for graph data understanding in online job recommendations,'' in \emph{AAAI}, vol.~38, no.~8, 2024, pp. 9178--9186.

\bibitem{tang2025grapharena}
J.~Tang, Q.~Zhang, Y.~Li, and J.~Li, ``Grapharena: Benchmarking large language models on graph computational problems,'' \emph{ICLR}, 2025.

\bibitem{zhang2024can}
Y.~Zhang, H.~Wang \emph{et~al.}, ``Can llm graph reasoning generalize beyond pattern memorization?'' in \emph{Findings of EMNLP}, 2024, pp. 2289--2305.

\bibitem{cao2024graphinsight}
Y.~Cao, S.~Han, Z.~Gao, Z.~Ding, X.~Xie, and S.~K. Zhou, ``Graphinsight: Unlocking insights in large language models for graph structure understanding,'' \emph{arXiv preprint arXiv:2409.03258}, 2024.

\bibitem{fatemi2024talk}
B.~Fatemi, J.~Halcrow, and B.~Perozzi, ``Talk like a graph: Encoding graphs for large language models,'' in \emph{ICLR}, 2024.

\bibitem{patil2024gorilla}
S.~G. Patil, T.~Zhang \emph{et~al.}, ``Gorilla: Large language model connected with massive apis,'' \emph{NeurIPS}, vol.~37, pp. 126\,544--126\,565, 2024.

\bibitem{zhang2023graph}
J.~Zhang, ``Graph-toolformer: To empower llms with graph reasoning ability via prompt augmented by chatgpt,'' \emph{arXiv preprint arXiv:2304.11116}, 2023.

\bibitem{chai2023graphllm}
Z.~Chai, T.~Zhang, L.~Wu, K.~Han, X.~Hu, X.~Huang, and Y.~Yang, ``Graphllm: Boosting graph reasoning ability of large language model,'' \emph{arXiv preprint arXiv:2310.05845}, 2023.

\bibitem{huang2024can2}
X.~Huang, K.~Han, Y.~Yang, D.~Bao, Q.~Tao, Z.~Chai, and Q.~Zhu, ``Can gnn be good adapter for llms?'' in \emph{WWW}, 2024, pp. 893--904.

\bibitem{duan2023simteg}
K.~Duan, Q.~Liu, T.-S. Chua, S.~Yan, W.~T. Ooi, Q.~Xie, and J.~He, ``Simteg: A frustratingly simple approach improves textual graph learning,'' \emph{arXiv preprint arXiv:2308.02565}, 2023.

\bibitem{chen2024label}
Z.~Chen, H.~Mao, H.~Wen, H.~Han, W.~Jin, H.~Zhang, H.~Liu, and J.~Tang, ``Label-free node classification on graphs with large language models (llms),'' in \emph{ICLR}, 2024.

\bibitem{chen2024exploring}
Z.~Chen, H.~Mao, H.~Li, W.~Jin, H.~Wen, X.~Wei, S.~Wang, D.~Yin, W.~Fan, H.~Liu \emph{et~al.}, ``Exploring the potential of large language models (llms) in learning on graphs,'' \emph{SIGKDD}, vol.~25, no.~2, pp. 42--61, 2024.

\bibitem{hu2023beyond}
Y.~Hu \emph{et~al.}, ``Beyond text: A deep dive into large language models' ability on understanding graph data,'' in \emph{NeurIPS Workshop}, 2023.

\bibitem{shu2024knowledge}
D.~Shu, T.~Chen, M.~Jin, C.~Zhang, M.~Du \emph{et~al.}, ``Knowledge graph large language model (kg-llm) for link prediction,'' in \emph{ACML}, 2024.

\bibitem{bi2024lpnl}
B.~Bi, S.~Liu, Y.~Wang \emph{et~al.}, ``Lpnl: Scalable link prediction with large language models,'' in \emph{Findings of ACL}, 2024, pp. 3615--3625.

\bibitem{ye2024language}
R.~Ye, C.~Zhang, R.~Wang, S.~Xu, and Y.~Zhang, ``Language is all a graph needs,'' in \emph{Findings of EACL}, 2024, pp. 1955--1973.

\bibitem{chen2024llaga}
R.~Chen, T.~Zhao, A.~K. JAISWAL, N.~Shah, and Z.~Wang, ``Llaga: Large language and graph assistant,'' in \emph{ICML}, 2024.

\bibitem{zhang2024graphtranslator}
M.~Zhang, M.~Sun, P.~Wang, S.~Fan, Y.~Mo, X.~Xu, H.~Liu, C.~Yang, and C.~Shi, ``Graphtranslator: Aligning graph model to large language model for open-ended tasks,'' in \emph{WWW}, 2024, pp. 1003--1014.

\bibitem{zhu2024efficient}
Y.~Zhu, Y.~Wang \emph{et~al.}, ``Efficient tuning and inference for large language models on textual graphs,'' in \emph{IJCAI}, 2024, pp. 5734--5742.

\bibitem{han2024pive}
J.~Han, N.~Collier, W.~Buntine, and E.~Shareghi, ``{P}i{V}e: Prompting with iterative verification improving graph-based generative capability of {LLM}s,'' in \emph{Findings of ACL}, Aug. 2024.

\bibitem{zhang2024extract}
B.~Zhang \emph{et~al.}, ``Extract, define, canonicalize: An llm-based framework for knowledge graph construction,'' in \emph{EMNLP}, 2024, pp. 9820--9836.

\bibitem{wen2024mindmap}
Y.~Wen, Z.~Wang, and J.~Sun, ``{M}ind{M}ap: Knowledge graph prompting sparks graph of thoughts in large language models,'' in \emph{ACL}, Aug. 2024.

\bibitem{vashishtha2023causal}
A.~Vashishtha, A.~G. Reddy \emph{et~al.}, ``Causal inference using llm-guided discovery,'' in \emph{AAAI Workshop}, 2023.

\bibitem{cao2023enhancing}
L.~Cao, ``Enhancing reasoning capabilities of large language models: A graph-based verification approach,'' \emph{arXiv:2308.09267}, 2023.

\bibitem{park2023graph}
J.~Park, A.~Patel, O.~Z. Khan, H.~J. Kim, and J.-K. Kim, ``Graph-guided reasoning for multi-hop question answering in large language models,'' \emph{arXiv preprint arXiv:2311.09762}, 2023.

\bibitem{da2024llm}
L.~Da, T.~Chen, L.~Cheng, and H.~Wei, ``Llm uncertainty quantification through directional entailment graph and claim level response augmentation,'' \emph{arXiv preprint arXiv:2407.00994}, 2024.

\bibitem{xu2024llm}
J.~Xu, Z.~Wu, M.~Lin, X.~Zhang, and S.~Wang, ``Llm and gnn are complementary: Distilling llm for multimodal graph learning,'' \emph{arXiv preprint arXiv:2406.01032}, 2024.

\bibitem{wen2023augmenting}
Z.~Wen \emph{et~al.}, ``Augmenting low-resource text classification with graph-grounded pre-training and prompting,'' in \emph{SIGIR}, 2023, pp. 506--516.

\bibitem{das2024modality}
D.~Das, I.~Gupta, J.~Srivastava, and D.~Kang, ``Which modality should i use-text, motif, or image?: Understanding graphs with large language models,'' in \emph{Findings of NAACL}, 2024, pp. 503--519.

\bibitem{sun2024head}
K.~Sun, Y.~Xu, H.~Zha, Y.~Liu, and X.~L. Dong, ``Head-to-tail: How knowledgeable are large language models (llms)? aka will llms replace knowledge graphs?'' in \emph{NAACL}, 2024, pp. 311--325.

\bibitem{he2024g}
X.~He, Y.~Tian, Y.~Sun, N.~Chawla, T.~Laurent \emph{et~al.}, ``G-retriever: Retrieval-augmented generation for textual graph understanding and question answering,'' \emph{NeurIPS}, vol.~37, pp. 132\,876--132\,907, 2024.

\bibitem{yang2024give}
L.~Yang, H.~Chen, Z.~Li, X.~Ding, and X.~Wu, ``Give us the facts: Enhancing large language models with knowledge graphs for fact-aware language modeling,'' \emph{TKDE}, 2024.

\bibitem{edge2024local}
D.~Edge, H.~Trinh, N.~Cheng, J.~Bradley \emph{et~al.}, ``From local to global: A graph rag approach to query-focused summarization,'' \emph{arXiv preprint arXiv:2404.16130}, 2024.

\bibitem{ma2024think}
S.~Ma, C.~Xu, X.~Jiang \emph{et~al.}, ``Think-on-graph 2.0: Deep and interpretable large language model reasoning with knowledge graph-guided retrieval,'' \emph{arXiv e-prints}, pp. arXiv--2407, 2024.

\bibitem{zhang2023autoalign}
R.~Zhang, Y.~Su, B.~D. Trisedya, X.~Zhao, M.~Yang, H.~Cheng, and J.~Qi, ``Autoalign: fully automatic and effective knowledge graph alignment enabled by large language models,'' \emph{TKDE}, 2023.

\bibitem{liu2024explore}
G.~Liu, Y.~Zhang, Y.~Li, and Q.~Yao, ``Explore then determine: A gnn-llm synergy framework for reasoning over knowledge graph,'' \emph{arXiv preprint arXiv:2406.01145}, 2024.

\bibitem{li2024llm}
Q.~Li, Z.~Chen, C.~Ji, S.~Jiang, and J.~Li, ``Llm-based multi-level knowledge generation for few-shot knowledge graph completion,'' in \emph{IJCAI}, vol. 271494703, 2024.

\bibitem{wang2024knowledge}
Y.~Wang, N.~Lipka, R.~A. Rossi, A.~Siu, R.~Zhang, and T.~Derr, ``Knowledge graph prompting for multi-document question answering,'' in \emph{AAAI}, vol.~38, no.~17, 2024, pp. 19\,206--19\,214.

\bibitem{shah2024improving}
M.~Shah and J.~Tian, ``Improving llm-based kgqa for multi-hop question answering with implicit reasoning in few-shot examples,'' in \emph{ACL KaLLM 2024}, 2024, pp. 125--135.

\bibitem{sehwag2024context}
U.~M. Sehwag, K.~Papasotiriou, J.~Vann, and S.~Ganesh, ``In-context learning with topological information for llm-based knowledge graph completion,'' in \emph{ICML Workshop}, 2024.

\bibitem{giarelis2024unified}
N.~Giarelis, C.~Mastrokostas, and N.~Karacapilidis, ``A unified llm-kg framework to assist fact-checking in public deliberation,'' in \emph{LREC-COLING Workshop}, 2024, pp. 13--19.

\bibitem{li2024optimus}
Z.~Li, Y.~Xie, R.~Shao, G.~Chen, D.~Jiang, and L.~Nie, ``Optimus-1: Hybrid multimodal memory empowered agents excel in long-horizon tasks,'' in \emph{NeurIPS}, 2024.

\bibitem{rana2023sayplan}
K.~Rana, J.~Haviland, S.~Garg, J.~Abou-Chakra, I.~Reid, and N.~Suenderhauf, ``Sayplan: Grounding large language models using 3d scene graphs for scalable robot task planning,'' in \emph{CoRL}, 2023.

\bibitem{chu2024llm}
Z.~Chu, Y.~Wang \emph{et~al.}, ``Llm-guided multi-view hypergraph learning for human-centric explainable recommendation,'' \emph{arXiv:2401.08217}, 2024.

\bibitem{abu2024knowledge}
H.~Abu-Rasheed, C.~Weber, and M.~Fathi, ``Knowledge graphs as context sources for llm-based explanations of learning recommendations,'' in \emph{EDUCON}, 2024, pp. 1--5.

\bibitem{ouyang2024modal}
K.~Ouyang, Y.~Liu, S.~Li \emph{et~al.}, ``Modal-adaptive knowledge-enhanced graph-based financial prediction from monetary policy conference calls with llm,'' in \emph{COLING Workshop}, 2024, pp. 59--69.

\bibitem{lu2024grace}
G.~Lu, X.~Ju, X.~Chen, W.~Pei, and Z.~Cai, ``Grace: Empowering llm-based software vulnerability detection with graph structure and in-context learning,'' \emph{JSS}, vol. 212, p. 112031, 2024.

\bibitem{li2024llm2}
W.~Li, Z.~Yu, Q.~She \emph{et~al.}, ``Llm-enhanced scene graph learning for household rearrangement,'' in \emph{SIGGRAPH}, 2024, pp. 1--11.

\bibitem{fang2020survey}
Y.~Fang, X.~Huang, L.~Qin, Y.~Zhang, W.~Zhang, R.~Cheng, and X.~Lin, ``A survey of community search over big graphs,'' \emph{VLDB Journal}, vol.~29, pp. 353--392, 2020.

\bibitem{eppstein1998finding}
D.~Eppstein, ``Finding the k shortest paths,'' \emph{SIAM Journal on computing}, vol.~28, no.~2, pp. 652--673, 1998.

\bibitem{wu2020comprehensive}
Z.~Wu, S.~Pan, F.~Chen \emph{et~al.}, ``A comprehensive survey on graph neural networks,'' \emph{TNNLS}, vol.~32, no.~1, pp. 4--24, 2020.

\bibitem{zhou2020graph}
J.~Zhou, G.~Cui, S.~Hu, Z.~Zhang \emph{et~al.}, ``Graph neural networks: A review of methods and applications,'' \emph{AI open}, vol.~1, pp. 57--81, 2020.

\bibitem{wu2022graph}
S.~Wu, F.~Sun \emph{et~al.}, ``Graph neural networks in recommender systems: a survey,'' \emph{ACM Computing Surveys}, vol.~55, no.~5, pp. 1--37, 2022.

\bibitem{zhao2023survey}
W.~X. Zhao, K.~Zhou, J.~Li, T.~Tang \emph{et~al.}, ``A survey of large language models,'' \emph{arXiv preprint arXiv:2303.18223}, 2023.

\bibitem{yang2024harnessing}
J.~Yang, H.~Jin, R.~Tang, X.~Han, Q.~Feng, H.~Jiang, S.~Zhong, B.~Yin, and X.~Hu, ``Harnessing the power of llms in practice: A survey on chatgpt and beyond,'' \emph{TKDD}, vol.~18, no.~6, pp. 1--32, 2024.

\bibitem{achiam2023gpt}
J.~Achiam, S.~Adler, S.~Agarwal, L.~Ahmad, I.~Akkaya, F.~L. Aleman, D.~Almeida, J.~Altenschmidt, S.~Altman, S.~Anadkat \emph{et~al.}, ``Gpt-4 technical report,'' \emph{arXiv preprint arXiv:2303.08774}, 2023.

\bibitem{chen2023token}
Y.~Chen, H.~Kang, V.~Zhai, L.~Li, R.~Singh, and B.~Raj, ``Token prediction as implicit classification to identify llm-generated text,'' in \emph{EMNLP}, 2023, pp. 13\,112--13\,120.

\bibitem{zhang2025gpt4roi}
S.~Zhang, P.~Sun, S.~Chen, M.~Xiao, W.~Shao, W.~Zhang, Y.~Liu, K.~Chen, and P.~Luo, ``Gpt4roi: Instruction tuning large language model on region-of-interest,'' in \emph{ECCV}, 2025, pp. 52--70.

\bibitem{wang2023far}
Y.~Wang, H.~Ivison, P.~Dasigi, J.~Hessel, T.~Khot, K.~Chandu, D.~Wadden, K.~MacMillan, N.~A. Smith, I.~Beltagy \emph{et~al.}, ``How far can camels go? exploring the state of instruction tuning on open resources,'' \emph{NeurIPS}, vol.~36, pp. 74\,764--74\,786, 2023.

\bibitem{francis2018cypher}
N.~Francis, A.~Green, and P.~e.~a. Guagliardo, ``Cypher: An evolving query language for property graphs,'' in \emph{SIGMOD}, 2018, pp. 1433--1445.

\bibitem{rodriguez2015gremlin}
M.~A. Rodriguez, ``The gremlin graph traversal machine and language (invited talk),'' in \emph{DBPL}, 2015, pp. 1--10.

\bibitem{perez2009semantics}
J.~P{\'e}rez, M.~Arenas, and C.~Gutierrez, ``Semantics and complexity of sparql,'' \emph{TODS}, vol.~34, no.~3, pp. 1--45, 2009.

\bibitem{wang2019rat}
B.~Wang, R.~Shin \emph{et~al.}, ``Rat-sql: Relation-aware schema encoding and linking for text-to-sql parsers,'' in \emph{ACL}, 2020, pp. 7567--7578.

\bibitem{shang2024survey}
W.~Shang and X.~Huang, ``A survey of large language models on generative graph analytics: Query, learning, and applications,'' \emph{arXiv preprint arXiv:2404.14809}, 2024.

\bibitem{liu2023pre}
P.~Liu, W.~Yuan, J.~Fu \emph{et~al.}, ``Pre-train, prompt, and predict: A systematic survey of prompting methods in natural language processing,'' \emph{ACM Computing Surveys}, vol.~55, no.~9, pp. 1--35, 2023.

\bibitem{schick2024toolformer}
T.~Schick, J.~Dwivedi-Yu, R.~Dess{\`\i} \emph{et~al.}, ``Toolformer: Language models can teach themselves to use tools,'' \emph{NeurIPS}, vol.~36, 2024.

\bibitem{wang2021gpt}
B.~Wang and A.~Komatsuzaki, ``Gpt-j-6b: A 6 billion parameter autoregressive language model,'' 2021.

\bibitem{touvron2023llama1}
H.~Touvron, T.~Lavril, G.~Izacard, X.~Martinet \emph{et~al.}, ``Llama: Open and efficient foundation language models,'' \emph{arXiv:2302.13971}, 2023.

\bibitem{wei2022chain}
J.~Wei, X.~Wang, D.~Schuurmans, M.~Bosma, F.~Xia, E.~Chi, Q.~V. Le, D.~Zhou \emph{et~al.}, ``Chain-of-thought prompting elicits reasoning in large language models,'' \emph{NeurIPS}, vol.~35, pp. 24\,824--24\,837, 2022.

\bibitem{wang2022self}
X.~Wang, J.~Wei, D.~Schuurmans \emph{et~al.}, ``Self-consistency improves chain of thought reasoning in language models,'' in \emph{ICLR}, 2023.

\bibitem{dong2024survey}
Q.~Dong, L.~Li, D.~Dai \emph{et~al.}, ``A survey on in-context learning,'' in \emph{EMNLP}, 2024, pp. 1107--1128.

\bibitem{schulman2017proximal}
J.~Schulman, F.~Wolski, P.~Dhariwal, A.~Radford, and O.~Klimov, ``Proximal policy optimization algorithms,'' \emph{arXiv:1707.06347}, 2017.

\bibitem{dong2023raft}
H.~Dong, W.~Xiong, D.~Goyal, Y.~Zhang, W.~Chow, R.~Pan, S.~Diao, J.~Zhang, K.~Shum, and T.~Zhang, ``Raft: Reward ranked finetuning for generative foundation model alignment,'' \emph{TMLR}, 2023.

\bibitem{song2024preference}
F.~Song, B.~Yu, M.~Li, H.~Yu, F.~Huang, Y.~Li, and H.~Wang, ``Preference ranking optimization for human alignment,'' in \emph{AAAI}, vol.~38, no.~17, 2024, pp. 18\,990--18\,998.

\bibitem{pan2024unifying}
S.~Pan, L.~Luo, Y.~Wang, C.~Chen, J.~Wang, and X.~Wu, ``Unifying large language models and knowledge graphs: A roadmap,'' \emph{TKDE}, 2024.

\bibitem{zhu2024llms}
Y.~Zhu, X.~Wang, and J.~e.~a. Chen, ``Llms for knowledge graph construction and reasoning: Recent capabilities and future opportunities,'' \emph{World Wide Web}, vol.~27, no.~5, p.~58, 2024.

\bibitem{tang2024graphgpt}
J.~Tang, Y.~Yang, W.~Wei, L.~Shi, L.~Su, S.~Cheng, D.~Yin, and C.~Huang, ``Graphgpt: Graph instruction tuning for large language models,'' in \emph{SIGIR}, 2024, pp. 491--500.

\bibitem{wu2024can}
X.~Wu, Y.~Shen, C.~Shan, K.~Song, S.~Wang, B.~Zhang, J.~Feng, H.~Cheng, W.~Chen, Y.~Xiong \emph{et~al.}, ``Can graph learning improve planning in llm-based agents?'' in \emph{NeurIPS}, 2024.

\bibitem{chen2023graph}
Z.~Chen, Z.~Jiang, F.~Yang, E.~Cho, X.~Fan, X.~Huang \emph{et~al.}, ``Graph meets llm: A novel approach to collaborative filtering for robust conversational understanding,'' in \emph{EMNLP}, 2023, pp. 811--819.

\bibitem{sun2023decoding}
C.~Sun, J.~Li, Y.~Fung, H.~Chan, T.~Abdelzaher \emph{et~al.}, ``Decoding the silent majority: Inducing belief augmented social graph with large language model for response forecasting,'' in \emph{EMNLP}, 2023, pp. 43--57.

\bibitem{peng2024chatgraph}
Y.~Peng, S.~Lin, Q.~Chen, S.~Wang, L.~Xu, X.~Ren, Y.~Li, and J.~Xu, ``Chatgraph: Chat with your graphs,'' in \emph{ICDE}, 2024, pp. 5445--5448.

\bibitem{hu2020open}
W.~Hu, M.~Fey, M.~Zitnik, Y.~Dong, H.~Ren, B.~Liu, M.~Catasta, and J.~Leskovec, ``Open graph benchmark: Datasets for machine learning on graphs,'' \emph{NeurIPS}, vol.~33, pp. 22\,118--22\,133, 2020.

\bibitem{mccallum2000automating}
A.~K. McCallum, K.~Nigam, J.~Rennie, and K.~Seymore, ``Automating the construction of internet portals with machine learning,'' \emph{Information Retrieval}, vol.~3, pp. 127--163, 2000.

\bibitem{giles1998citeseer}
C.~L. Giles, K.~D. Bollacker, and S.~Lawrence, ``Citeseer: An automatic citation indexing system,'' in \emph{Proceedings of the third ACM conference on Digital libraries}, 1998, pp. 89--98.

\bibitem{tang2008arnetminer}
J.~Tang, J.~Zhang, L.~Yao \emph{et~al.}, ``Arnetminer: extraction and mining of academic social networks,'' in \emph{SIGKDD}, 2008, pp. 990--998.

\bibitem{zhang2018variational}
Y.~Zhang, H.~Dai, Z.~Kozareva \emph{et~al.}, ``Variational reasoning for question answering with knowledge graph,'' in \emph{AAAI}, vol.~32, no.~1, 2018.

\bibitem{wang2021kepler}
X.~Wang, T.~Gao, Z.~Zhu, Z.~Zhang, Z.~Liu, J.~Li, and J.~Tang, ``Kepler: A unified model for knowledge embedding and pre-trained language representation,'' \emph{ACL}, vol.~9, pp. 176--194, 2021.

\bibitem{borgwardt2005protein}
K.~M. Borgwardt, C.~S. Ong, S.~Sch{\"o}nauer, S.~Vishwanathan, A.~J. Smola, and H.-P. Kriegel, ``Protein function prediction via graph kernels,'' \emph{Bioinformatics}, vol.~21, no. suppl\_1, pp. i47--i56, 2005.

\bibitem{debnath1991structure}
A.~K. Debnath \emph{et~al.}, ``Structure-activity relationship of mutagenic aromatic and heteroaromatic nitro compounds. correlation with molecular orbital energies and hydrophobicity,'' \emph{Journal of Medicinal Chemistry}, vol.~34, no.~2, pp. 786--797, 1991.

\bibitem{wale2008comparison}
N.~Wale, I.~A. Watson, and G.~Karypis, ``Comparison of descriptor spaces for chemical compound retrieval and classification,'' \emph{Knowledge and Information Systems}, vol.~14, pp. 347--375, 2008.

\bibitem{toivonen2003statistical}
H.~Toivonen, A.~Srinivasan, R.~D. King, S.~Kramer, and C.~Helma, ``Statistical evaluation of the predictive toxicology challenge 2000--2001,'' \emph{Bioinformatics}, vol.~19, no.~10, pp. 1183--1193, 2003.

\bibitem{kong2013inferring}
X.~Kong, J.~Zhang, and P.~S. Yu, ``Inferring anchor links across multiple heterogeneous social networks,'' in \emph{CIKM}, 2013, pp. 179--188.

\bibitem{yih2016value}
W.-t. Yih, M.~Richardson \emph{et~al.}, ``The value of semantic parse labeling for knowledge base question answering,'' in \emph{ACL}, 2016, pp. 201--206.

\bibitem{talmor2018web}
A.~Talmor and J.~Berant, ``The web as a knowledge-base for answering complex questions,'' in \emph{NAACL}, 2018, pp. 641--651.

\bibitem{gu2021beyond}
Y.~Gu, S.~Kase, M.~Vanni, B.~Sadler, P.~Liang, X.~Yan, and Y.~Su, ``Beyond iid: three levels of generalization for question answering on knowledge bases,'' in \emph{WWW}.\hskip 1em plus 0.5em minus 0.4em\relax ACM, 2021, pp. 3477--3488.

\bibitem{perevalov2022qald}
A.~Perevalov, D.~Diefenbach, R.~Usbeck, and A.~Both, ``Qald-9-plus: A multilingual dataset for question answering over dbpedia and wikidata translated by native speakers,'' in \emph{ICSC}, 2022, pp. 229--234.

\bibitem{wan2018item}
M.~Wan and J.~McAuley, ``Item recommendation on monotonic behavior chains,'' in \emph{RecSys}, 2018, pp. 86--94.

\bibitem{ni2019justifying}
J.~Ni, J.~Li \emph{et~al.}, ``Justifying recommendations using distantly-labeled reviews and fine-grained aspects,'' in \emph{EMNLP}, 2019, pp. 188--197.

\bibitem{bojchevski2018deep}
A.~Bojchevski and S.~G{\"u}nnemann, ``Deep gaussian embedding of graphs: Unsupervised inductive learning via ranking,'' in \emph{ICLR}, 2018.

\bibitem{shchur2018pitfalls}
O.~Shchur, M.~Mumme, A.~Bojchevski, and S.~G{\"u}nnemann, ``Pitfalls of graph neural network evaluation,'' \emph{arXiv:1811.05868}, 2018.

\bibitem{rossi2015network}
R.~Rossi and N.~Ahmed, ``The network data repository with interactive graph analytics and visualization,'' in \emph{AAAI}, vol.~29, no.~1, 2015.

\bibitem{huang2020analysis}
H.~Huang \emph{et~al.}, ``An analysis framework of research frontiers based on the large-scale open academic graph,'' \emph{ASIST}, vol.~57, p. e307, 2020.

\bibitem{cobbe2021training}
K.~Cobbe, V.~Kosaraju, M.~Bavarian \emph{et~al.}, ``Training verifiers to solve math word problems,'' \emph{arXiv preprint arXiv:2110.14168}, 2021.

\bibitem{patel2021nlp}
A.~Patel, S.~Bhattamishra, and N.~Goyal, ``Are nlp models really able to solve simple math word problems?'' in \emph{NAACL}, 2021, pp. 2080--2094.

\bibitem{han2024folio}
S.~Han, H.~Schoelkopf, Y.~Zhao \emph{et~al.}, ``Folio: Natural language reasoning with first-order logic,'' in \emph{EMNLP}, 2024, pp. 22\,017--22\,031.

\bibitem{scutari2015bayesian}
M.~Scutari \emph{et~al.}, ``Bayesian networks with examples in r,'' 2015.

\bibitem{xia2022medconqa}
F.~Xia, B.~Li \emph{et~al.}, ``{M}ed{C}on{QA}: Medical conversational question answering system based on knowledge graphs,'' in \emph{EMNLP}, 2022.

\bibitem{li2023chatdoctor}
Y.~Li, Z.~Li, K.~Zhang, R.~Dan, S.~Jiang, and Y.~Zhang, ``Chatdoctor: A medical chat model fine-tuned on a large language model meta-ai (llama) using medical domain knowledge,'' \emph{Cureus}, vol.~15, no.~6, 2023.

\bibitem{zhang2023effect}
Y.~Zhang \emph{et~al.}, ``The effect of metadata on scientific literature tagging: A cross-field cross-model study,'' in \emph{WWW}, 2023, pp. 1626--1637.

\bibitem{johnson2016mimic}
A.~E. Johnson, T.~J. Pollard, L.~Shen \emph{et~al.}, ``Mimic-iii, a freely accessible critical care database,'' \emph{Scientific data}, vol.~3, no.~1, pp. 1--9, 2016.

\bibitem{bollacker2008freebase}
K.~Bollacker, C.~Evans, P.~Paritosh, T.~Sturge, and J.~Taylor, ``Freebase: a collaboratively created graph database for structuring human knowledge,'' in \emph{SIGMOD}, 2008, pp. 1247--1250.

\bibitem{toutanova2015observed}
K.~Toutanova and D.~Chen, ``Observed versus latent features for knowledge base and text inference,'' in \emph{ACL Workshop}, 2015, pp. 57--66.

\bibitem{lin2004rouge}
C.-Y. Lin, ``Rouge: A package for automatic evaluation of summaries,'' in \emph{Text summarization branches out}, 2004, pp. 74--81.

\bibitem{papineni2002bleu}
K.~Papineni, S.~Roukos \emph{et~al.}, ``Bleu: a method for automatic evaluation of machine translation,'' in \emph{ACL}, 2002, pp. 311--318.

\bibitem{goutte2005probabilistic}
C.~Goutte \emph{et~al.}, ``A probabilistic interpretation of precision, recall and f-score, with implication for evaluation,'' in \emph{ECIR}, 2005, pp. 345--359.

\bibitem{kleinberg1999web}
J.~M. Kleinberg, R.~Kumar \emph{et~al.}, ``The web as a graph: Measurements, models, and methods,'' in \emph{COCOON}, 1999, pp. 1--17.

\bibitem{iacus2012causal}
S.~M. Iacus, G.~King \emph{et~al.}, ``Causal inference without balance checking: Coarsened exact matching,'' \emph{Political analysis}, vol.~20, pp. 1--24, 2012.

\bibitem{zhao2023gimlet}
H.~Zhao, S.~Liu, M.~Chang, H.~Xu, J.~Fu, Z.~Deng, L.~Kong, and Q.~Liu, ``Gimlet: A unified graph-text model for instruction-based molecule zero-shot learning,'' \emph{NeurIPS}, vol.~36, pp. 5850--5887, 2023.

\bibitem{mao2024position}
H.~Mao, Z.~Chen, and W.~e.~a. Tang, ``Position: Graph foundation models are already here,'' in \emph{ICML}, 2024.

\bibitem{garza2024priv}
L.~Garza, L.~Elluri \emph{et~al.}, ``Privcomp-kg: Leveraging knowledge graph and large language models for privacy policy compliance verification,'' \emph{arXiv:2404.19744}, 2024.

\bibitem{guan2024graph}
F.~Guan, T.~Zhu, W.~Zhou, and K.-K.~R. Choo, ``Graph neural networks: a survey on the links between privacy and security,'' \emph{Artificial Intelligence Review}, vol.~57, no.~2, p.~40, 2024.

\bibitem{liu2025tigervector}
S.~Liu, Z.~Zeng, L.~Chen \emph{et~al.}, ``Tigervector: Supporting vector search in graph databases for advanced rags,'' in \emph{ICMD}, 2025, pp. 553--565.

\bibitem{khan2025graph}
A.~Khan, X.~Ke \emph{et~al.}, ``Graph data management and graph machine learning: Synergies and opportunities,'' \emph{arXiv:2502.00529}, 2025.

\end{thebibliography}


 



\vfill

\end{document}